\documentclass[iicol,pdflatex,sn-mathphys-ay]{sn-jnl}

\usepackage{graphicx}%
\usepackage{multirow}%
\usepackage{amsmath,amssymb,amsfonts}%
\usepackage{amsthm}%
\usepackage{mathrsfs}%
\usepackage[title]{appendix}%
\usepackage{xcolor}%
\usepackage{textcomp}%
\usepackage{manyfoot}%
\usepackage{booktabs}%
\usepackage{algorithm}%
\usepackage{algorithmicx}%
\usepackage{algpseudocode}%
\usepackage{listings}%
\usepackage{ragged2e}

\theoremstyle{thmstyleone}%
\theoremstyle{thmstyletwo}%

\theoremstyle{thmstylethree}%

\begin{document}

\title[A-TDOM: Active TDOM via On-the-Fly 3DGS]{A-TDOM: Active TDOM via On-the-Fly 3DGS}



\author[1]{\fnm{Yiwei} \sur{Xu}}\email{xywjohn\_sgg2020@whu.edu.cn}
\author[1]{\fnm{Xiang} \sur{Wang}}\email{xiangwang@whu.edu.cn}
\author[1]{\fnm{Yifei} \sur{Yu}}\email{yfyu2020@whu.edu.cn}
\author[1]{\fnm{Wentian} \sur{Gan}}\email{gwt2019@whu.edu.cn}
\author[2]{\fnm{Luca} \sur{Morelli}}\email{lmorelli@fbk.eu}
\author[2]{\fnm{Giulio} \sur{Perda}}\email{gperda@fbk.eu }
\author*[1]{\fnm{Xin} \sur{Wang}}\email{xwang@sgg.whu.edu.cn}  
\author[1]{\fnm{Zongqian} \sur{Zhan}}\email{zqzhan@sgg.whu.edu.cn}
\author[2]{\fnm{Fabio} \sur{Remondino}}\email{remondino@fbk.eu}

\affil[1]{\orgdiv{School of Geodesy and Geomatics}, 
\orgname{Wuhan University}, 
\orgaddress{\city{Wuhan}, \country{China}}}

\affil[2]{\orgdiv{Fondazione Bruno Kessler}, 
\orgname{3D Optical Metrology Unit}, 
\orgaddress{\city{Trento}, \country{Italy}}}


\abstract
{
True Digital Orthophoto Map (TDOM), a 2D objective representation of the Earth's surface, is an essential geospatial product widely used in urban management, city planning, land surveying, and related applications. However, traditional TDOM generation typically relies on a complex offline photogrammetric pipeline, leading to substantial latency and making it unsuitable for time-critical or real-time scenarios. Moreover, the quality of TDOM may deteriorate due to inaccurate camera poses, imperfect Digital Surface Model (DSM), and incorrect occlusions detection. To address these challenges, this work introduces \textbf{A-TDOM}, a near real-time TDOM generation method built upon On-the-Fly 3DGS (3D Gaussian Splatting) optimization. As each incoming image arrives, its pose and sparse point cloud are computed via On-the-Fly SfM. Newly observed regions are then incrementally reconstructed as additional 3D Gaussians are inserted using a Delaunay triangulated Gaussian sampling and integration and are further optimized via adaptive training iterations and learning rate, especially in previously unseen or coarsely modeled areas. With orthogonal splatting integrated into the rendering pipeline, A-TDOM can actively produce updated TDOM outputs immediately after each 3DGS update. Code is now available at \href{ https://github.com/xywjohn/A-TDOM}{ https://github.com/xywjohn/A-TDOM}
}




\maketitle

\section{Introduction}\label{Introduction}

Aerial True Digital Orthophoto Map (TDOM) features high resolution and rich texture information, making it widely applied in various fields (e.g., digital twin, city planning, etc.). Traditionally, the generation of a TDOM begins with aerial triangulation to determine image orientation, followed by a complex post-processing procedure to derive the corresponding Digital Surface Model (DSM). Compared to the Digital Orthophoto Map (DOM), TDOM generation requires both orthorectification and occlusion detection (e.g., Z-buffering \citep{uchida2004triangle}) based on the DSM, both of which are computationally intensive operations.

\begin{figure*}[t]
    \centering
    \includegraphics[trim={0 0 0 0},clip, width=\textwidth]{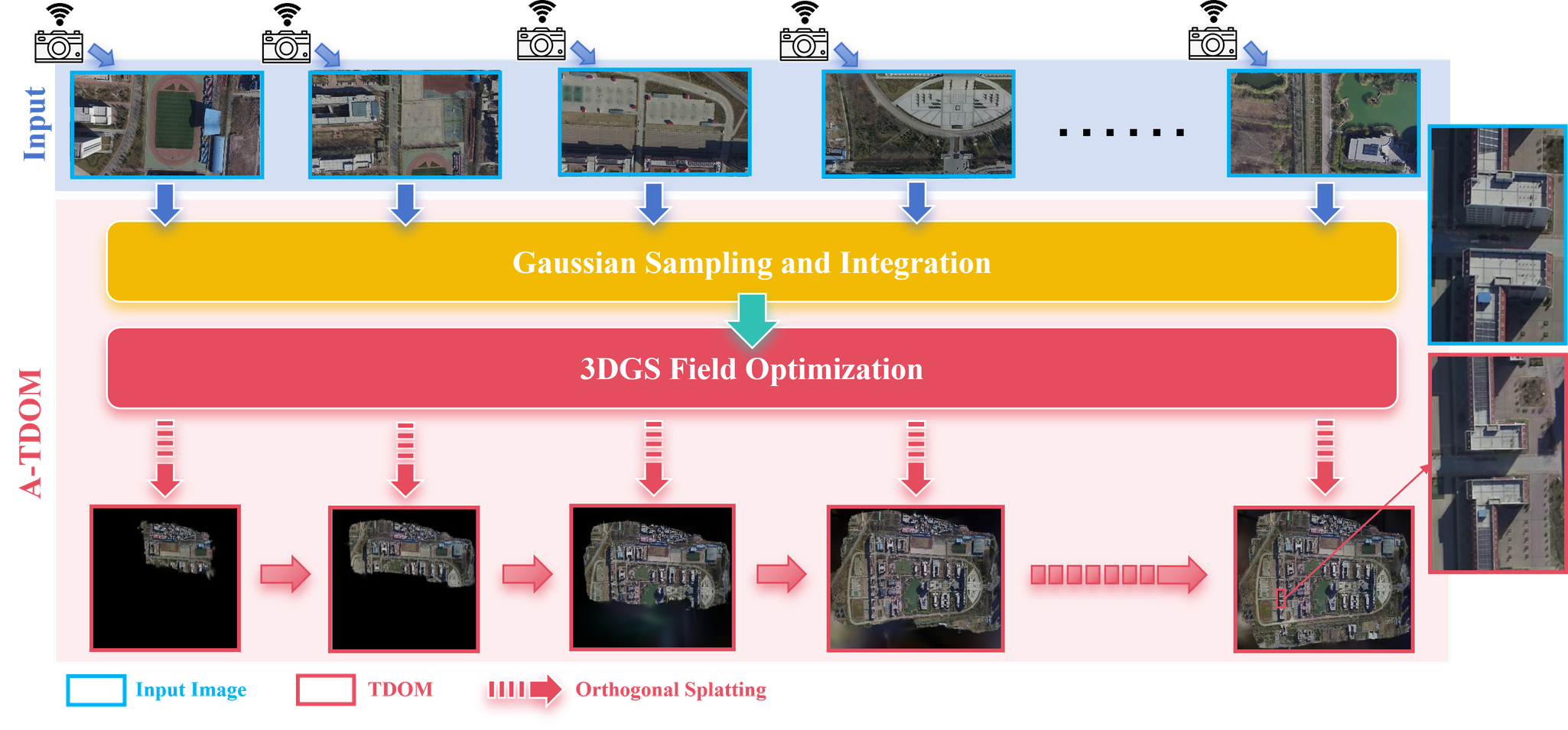}
    \caption{Our A-TDOM. For each input image, it pose and corresponding new point cloud are estimated and the current 3DGS field is updated via the proposed Gaussians Sampling and Integration method, followed by the 3DGS field optimization. Subsequently, TDOM is generated using the Orthogonal Splatting technique.}
    \label{fig:0}
\end{figure*}

In recent years, the emergence of 3D Gaussian Splatting (3DGS) \citep{3DGS} has introduced a novel paradigm for large-scale scene reconstruction and TDOM generation. 3DGS-based methods express a scene by optimizing a set of Gaussians, offering a more streamlined pipeline compared to traditional photogrammetric workflows, while achieving notable advantages in both reconstruction efficiency and rendering quality. To enable TDOM generation, several studies have first focused on large-scale scene reconstruction using 3DGS. For example, CityGaussian \citep{CityGaussian} achieves city-scale 3DGS model training through tile-based parallel optimization and a Level of Detail (LoD) algorithm, whereas RetinaGS \citep{Retinags} adopts distributed training with sub-models and improves the Gaussians initialization strategy, both effectively achieving large-scale 3DGS model training with improved quality, thereby providing a solid foundation for 3DGS-based TDOM generation. Subsequently, to enable TDOM generation based on 3DGS field, Tortho-Gaussian \citep{Tortho-gaussian} and Ortho-3DGS \citep{Ortho-3DGS} have employed an Orthogonal Splatting approach. By performing orthographic projection on the mean and covariance of each Gaussian, any 3DGS field can be orthographically rendered from arbitrary views. Compared with the traditional pipeline, 3DGS-based methods are capable of generating high-quality TDOM without DSM and occlusion detection, representing substantial advancement over conventional workflows.

However, most existing methods still employ offline processing pipeline after the acquisition of all images, limiting the potential of real-time applications. There are few works for real-time DOM generation \citep{wei2008real, hinzmann2017mapping, du2025real, ARTEMIS, zhou2019real}, but not for TDOM. Additionally, although SLAM-based methods (e.g., Gaussian Splatting SLAM \citep{GS-SLAM}, RTG-SLAM \citep{RTG-SLAM}) have achieved real-time 3DGS training, they are limited to processing low-resolution images captured in small-scale scenes and require continuously input frames with strong spatiotemporal continuity, which restricts their applicability to large-scale and unconstrained 3D reconstruction and rendering scenarios. To address these limitations, this paper presents A-TDOM, a workflow which enables simultaneous image acquisition and active TDOM generation based on On-the-Fly 3DGS, as shown in Fig. \ref{fig:0}. Compared to the offline 3DGS training solutions, On-the-Fly 3DGS deals with a continuously expanding dynamic scenario. Therefore, integrating new Gaussians to the current 3DGS field in previously unseen and coarsely reconstructed regions becomes crucial. To overcome this challenge, this paper employs a Gaussians Sampling and Integration method. For each input image, it pose and corresponding new point cloud are estimated online by On-the-Fly SfM \citep{On-the-Fly-SfM}. The generated point cloud is reprojected onto the image plane, where the visible 2D points are used to construct a Delaunay triangulation. The current 3DGS field is then used to render the newly input image, whose rendering quality is analyzed to identify potentially previously unseen and coarsely reconstructed regions within the image coverage. Subsequently, new 2D points are sampled adaptively within these regions, and their 3D coordinates and color attributes are computed based on the geometry of the corresponding Delaunay triangles. All 3D points are then initialized as new Gaussians and integrated into the current 3DGS field, enabling a continuous, adaptive 3DGS expansion as the image sequence grows.

To achieve real-time/near real-time optimization of 3DGS, the following strategies are adopted in this work. First, the original 3DGS densification strategy \citep{3DGS} is disabled, and new Gaussians are introduced solely through the proposed Gaussian Sampling and Integration method. Meanwhile, the original Gaussian pruning strategy is retained to eliminate redundant Gaussians, thereby maintaining control over the number of Gaussians and improving optimization efficiency while reducing memory consumption. Furthermore, to accelerate optimization in previously unseen regions, A-TDOM allocates more training iterations to newly input images during 3DGS optimization. Meanwhile, learning rates are assigned per image, meaning that newly input and coarsely reconstructed images are assigned higher learning rates to enable rapid optimization, while images with well-reconstructed regions are assigned lower learning rates to achieve stable and continuous refinement. Experimental results presented in Sect.\ref{Experiments} demonstrate that A-TDOM can complete the training of a newly input image within 1-2 seconds, successfully achieving near real-time optimization of 3DGS. Upon completing the training of each newly input image, the current 3DGS field is orthographically rendered using Orthogonal Splatting, producing a progressively updated TDOM within 20-30 milliseconds.

In summary, this work makes the following contributions:

\begin{itemize}
    \item A-TDOM, a worflow to enable robust online active TDOM generation without DSM and occlusion detection. To the best of our knowledge, it is the first attempt to generate TDOM in near real-time. 
    \item A near real-time 3DGS optimization, based on Gaussians Sampling and Integration method to novely update the current 3DGS field for each new image. Subsequently, several adaptive optimization strategies are proposed to accelerate the optimization for previously unseen regions.
    \item An innovative approach, based on projection matrix modification, to enable orthographic splatting and obtain TDOM from 3DGS.
    
\end{itemize}

\section{Related Works}\label{Related_Works}

In this section, three relevant studies are reviewed including novel view synthesis, real-time 3DGS training, generation of DOM and TDOM.

\subsection{Novel View Synthesis}

Novel View Synthesis (NVS) has been an active research topic in both computer vision and computer graphics for many years \citep{Survey_On_3DGS}. Traditionally, various approaches have been developed to achieve NVS. Light Field Rendering employs light field cameras to capture light field data, enabling novel view synthesis by interpolation \citep{LFR,ULF}. The Lumigraph method \citep{lumigraph, Unstructured_lumigraph} improves Light Field Rendering with greater flexibility and broader applicability. View Interpolation \citep{View_interpolation} utilizes multi-view images to interpolate among existing images while still employing a traditional photogrammetry method to generate a mesh model of the scene and render images from arbitrary viewpoints using back-projection. 

In recent years, the advent of Neural Radiance Fields (NeRF) \citep{NeRF} and 3DGS \citep{3DGS} has introduced novel paradigms for NVS. NeRF trains a multi-layer perceptron (MLP) to implicitly represent a 3D scene, achieving high-quality novel view rendering. Recent studies have enhanced its performance from multiple perspectives. Instant-NGP \citep{Instant-NGP} employs a multiresolution hash encoding to replace the fully connected input layer, significantly accelerating both training and rendering. NeRF-W \citep{NeRF-W} and NeRF++ \citep{NeRF++} extend NeRF to handle crowd-sourced image collections and large-scale outdoor scenes, thereby broadening its applicability. Meanwhile, Block-NeRF \citep{Block-NeRF}, Mega-NeRF \citep{Mega-NeRF} and Mega-NeRF++ \citep{4_nerf_Mega-nerf-plus} adopt a blocking strategy that divides a large-scale and complex scene into multiple sub-blocks for parallel training, successfully enabling the reconstruction of extensive urban-scale environments. 

Compared with learning-based implicit scene representation methods, 3DGS adopts explicit Gaussians to represent the scene, providing geometrically meaningful and more interpretable representations while enabling fast training and real-time rendering. Subsequently, numerous follow-up works further improve the efficiency and the rendering quality of 3DGS via various strategies. In terms of efficiency, AdR-Gaussian \citep{AdR-Gaussian} and EfficientGS \citep{EfficientGS} adopt different pruning strategies to remove low-contribution Gaussians, thereby reducing the number of Gaussians and accelerating both training and rendering. Taming-3DGS \citep{Taming-3DGS} improves the back-propagation process by adopting a Gaussian-based parallel strategy when accumulating gradients for each Gaussian, significantly enhancing training efficiency. \cite{papantonakis2024reducing} employ a half-float representation and a codebook-based quantization strategy, successfully reducing storage requirements and improving efficiency. In terms of rendering quality, \cite{hahlbohm2025efficient} maximize multi-view coherence and perspective accuracy through a perspective-correct evaluation strategy, thereby improving rendering consistency. \cite{CLIPGaussian} incorporate semantic information during training and introduce a 3D spatial consistency loss, enabling robust reconstruction under low-quality or blurred image conditions. SparseGS \citep{SparseGS} leverages depth estimation and geometric priors to strengthen geometric constraints in sparse-view scenarios, achieving higher rendering quality from limited input views. Furthermore, since most existing methods focus on small-scale scene representation, several works have explored large-scale 3DGS training and applications. RetinaGS \citep{Retinags} and DOGS \citep{DOGS} adopt image tiling to support 3DGS training for large-scale scenes, while PGSR \citep{PGSR} and PG-SAG \citep{PG-SAG} further enable surface reconstruction of urban scenes based on the 3DGS framework. However, these methods still operate within an offline processing pipeline after the acquisition of all images, thus limiting their potential for real-time applications, even though they have achieved efficient 3DGS training and rendering.

\subsection{3DGS Real-time Training}

To achieve real-time performance, most existing studies (e.g., VPGS-SLAM \citep{VPGS-SLAM}, RTG-SLAM \citep{RTG-SLAM}, and Compact-3DGS \citep{Compact-3DGS}) have explored 3DGS via the integration with SLAM. They select keyframes from continuous video streams and assign them with initial camera poses. Relying on depth priors, new Gaussians are then added to the current 3DGS field and jointly optimized with the corresponding camera poses during training. Furthermore, some SLAM-based approaches (e.g., Gaussian Splatting SLAM \citep{GS-SLAM}, WildGS-SLAM \citep{WildGS-SLAM}) have also achieved real-time 3DGS training without relying on depth priors, indicating their broader applicability. However, SLAM-based methods are generally designed for video frames in small-scale scenes and struggle to handle high-resolution images that are captured in large-scale scenes and lack spatiotemporal continuity (e.g. photogrammetric image dataset).

To overcome these limitations, several studies have explored real-time 3DGS training beyond SLAM-based frameworks. On-the-Fly NVS \citep{On-the-Fly-NVS} samples new Gaussians guided by monocular depth estimated from Depth-Anything-V2 \citep{Depth-Anything-V2}. Moreover, it refines the Gaussian initialization strategy and introduces a Scalable Incremental Gaussian Construction scheme to accelerate optimization and reduce memory consumption, achieving efficient online training under constrained computational resources. Recently, Gaussian On-the-Fly Splatting \citep{On-the-Fly-GS} investigates a new Gaussian initialization and integration based on continuously updated sparse point clouds. Benefiting from a hierarchical image weighting strategy, the learning rates and training iterations are assigned adaptively to different registered images, ensuring rapid optimization for previously unseen regions.

Although these methods have demonstrated the feasibility of real-time 3DGS training on photogrammetric image datasets, they still suffer from reduced training efficiency and significantly high computational cost when dealing with large-scale and complex scenes. In particular, for real-time or near-real-time TDOM generation, maintaining rendering quality while preserving real-time performance becomes even more critical.

\subsection{Generation of DOM and TDOM}

Over the past decades, the generation of DOM and TDOM has been extensively studied \citep{hood1989image, Review_DOM}, which typically relies on offline processing \citep{dewitt2000elements, lin2018research, zhao2022novel}. Traditionally, \cite{dewitt2000elements} provide a comprehensive introduction to the generation of DOM based on conventional photogrammetric workflows. More recent studies \citep{lin2018research, zhao2022novel, yuan2023fully} have focused on improving specific stages within the traditional DOM generation pipeline. For example, \cite{yuan2023fully} propose a pixel-by-pixel digital differential rectification method for DOM generation, which improves the geometric accuracy of DOMs while maintaining computational efficiency. In addition, learning-based approaches have been explored for DOM generation. \cite{li2019topological} employ neural networks to enhance the reconstruction accuracy of building edges, thereby improving the quality of DOM. Moreover, \cite{Review_DOM} have attempted to generate DOM using NeRF and verified the feasibility of this concept through a comparative analysis with traditional DOM generation workflow. Nevertheless, some works have successfully demonstrated real-time generation of DOM. \cite{wei2008real} perform video image matching based on control points and employ a fast algorithm for DOM generation, successfully achieving real-time DOM generation. \cite{hinzmann2017mapping} propose a real-time mapping framework for UAVs, which employs a cost-efficient sensor setup for pose estimation and optimizes memory usage and processing performance across diverse terrains, achieving robust real-time mapping and DOM generation. \cite{du2025real} introduce a learning-based approach for real-time DOM generation. It estimates pose information for each image based on deep features and employs optimal orthogonality values and multiband blending algorithms, generating DOM in a real-time manner. \cite{ARTEMIS} propose an efficient feature matching strategy combined with adaptive bundle adjustment, enabling real-time generation of DOM. \cite{zhou2019real} present an FPGA-based rapid ortho-rectification strategy, also enabling real-time DOM generation.

In contrast to the investigations in real-time DOM generation, relevant works on real-time TDOM generation remain relatively limited. To the best of our knowledge, existing methods have only realized TDOM generation in an offline manner. Traditionally, \cite{uchida2004triangle} generate a DSM through conventional photogrammetric workflow and perform orthorectification and occlusion detection to produce TDOM. NeRF-based methods \citep{NeRFOrtho, wei2024nerftdom} train a MLP in an offline manner and employ an ortho-volume rendering strategy for TDOM generation. 3DGS-based methods \citep{Tortho-gaussian, Ortho-3DGS} also optimize the 3DGS field through offline workflow and perform orthographic splatting from specific viewpoints to obtain TDOM. To address these limitations, the proposed A-TDOM achieves an efficient and robust real-time training of 3DGS, enabling active and continuous TDOM generation with orthogonal splatting.

\section{Preliminaries}\label{Preliminaries}

\subsection{3D Gaussian Splatting}

3DGS is denoted as an explicit 3D representation that utilizes a large number of Gaussians to express a real 3D scene \citep{3DGS}. Each Gaussian contains both geometric and material properties. The geometric properties characterize the spatial position and shape of each Gaussian, whereas the material properties mainly describe its color distribution and opacity.

\textbf{Geometric properties.} For each Gaussian, with a given rotation matrix $R$ and the scaling matrix $S$, the Gaussian centered at $X_0$ is defined in world space as follows:

$$
G(X)=e^{-\frac{1}{2}\left(X-X_{0}\right)^{T} \boldsymbol{\Sigma}^{-1}\left(X-X_{0}\right)} \eqno{(1)}
$$
where $X$ is the coordinate of a sampling point and the covariance matrix of the Gaussian $\boldsymbol{\Sigma}$ is calculated as $\boldsymbol{\Sigma}= RSS^{T}R^{T}$. 

\textbf{Material properties.} 3DGS employs spherical harmonics (SH) to represent the color distribution of each Gaussian, and $\sigma$ is the opacity of the Gaussian. Before calculating the color of one specific pixel, 3DGS estimates the splatting opacity $\alpha$ based on the Gaussian-pixel 2D distance $x$ and the projected covariance matrix $\boldsymbol{\Sigma}'$: 

$$
\alpha =\sigma e^{ -\frac{1}{2}x^{T}\boldsymbol{\Sigma}^{'-1}x} \eqno{(2)}
$$

For a set of Gaussians $G_N$ which are associated with a specific pixel, if the color of each Gaussian $i$ is denoted as $c_{i}$ and its splatting opacity as $\alpha_{i}$, the color of the corresponding pixel $c_\text{pixel}$ can be integrated as follows:

$$
c_\text{pixel}=\sum_{i=1}^{G_N}\left ( c_{i}\alpha_{i}\prod_{j=1}^{i-1}\left ( 1-\alpha_{j}   \right )    \right ) \eqno{(3)}
$$

\begin{figure*}[t]
    \centering
    \includegraphics[trim={0.65cm 0 0 0},clip, width=1.0\textwidth]{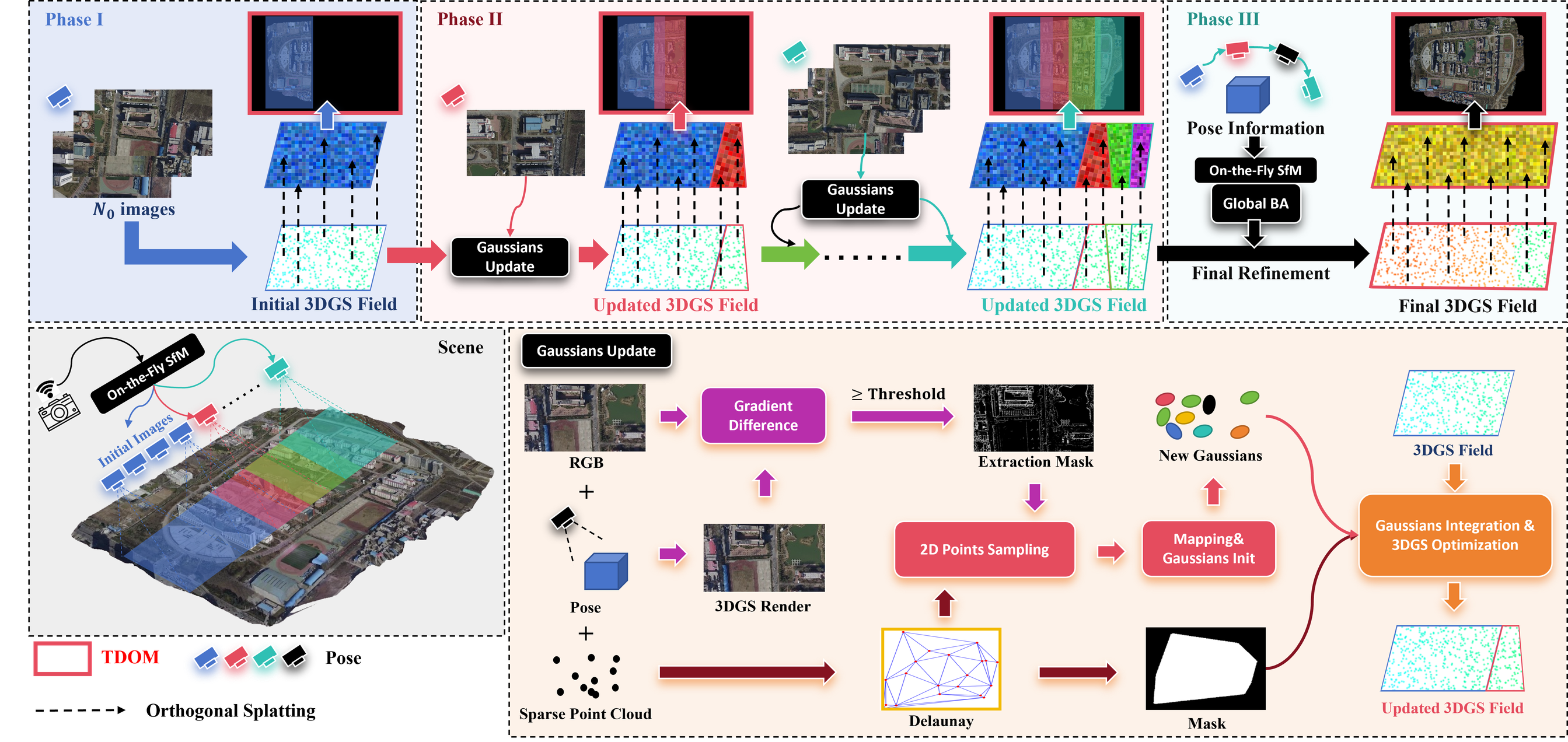}
    \caption{The workflow of our A-TDOM. After an initial 3DGS field, for each new image, a Delaunay triangulation is applied on the 3D-2D reprojections to create a mask for 3DGS field updating and optimization. The TDOM is generated after each 3DGS field update via orthogonal splatting.}
    \label{fig:1}
\end{figure*}

During a novel view rendering, 3DGS assigns a dedicated thread to each pixel, enabling parallel computation of the color of each pixel to facilitate high-quality, real-time rendering.

\subsection{On-the-Fly SfM}

Traditional photogrammetric workflows and SfM frameworks (e.g., COLMAP \citep{COLMAP}) typically perform pose estimation and sparse point cloud generation after the aquisition of all images. Such a sequential processing scheme makes them unsuitable for A-TDOM, which demands concurrent scene reconstruction and TDOM generation during image acquisition. On the other hands, SLAM-based methods are designed to process continuous video streams, where the input frames exhibit strong spatiotemporal consistency, limiting their applicability to unconstrained 3D reconstruction and rendering scenarios. To overcome this limitation, we rely on On-the-Fly SfMv2 \citep{On-the-Fly-SfM} (See more at \href{ https://yifeiyu225.github.io/on-the-flySfMv2.github.io/}{ https://yifeiyu225.github.io/on-the-flySfMv2.github.io/}), which achieves near real-time pose estimation and sparse point cloud during arbitrary image capturing, for our subsequent implementation. For each newly input image, the On-the-Fly SfM extracts global features using a pre-trained CNN and performs online feature matching via HNSW \citep{HNSW}, thereby enabling real-time updates of the image matching matrix. During pose estimation and sparse point cloud generation, On-the-Fly SfM employs a Hierarchical Weighted Local Bundle Adjustment that only newly input image and it neighboring connected images are incorporated into Bundle Adjustment (BA). Compared with COLMAP, which frequently performs global BA whose computational cost increases significantly as the number of registered images increases, Hierarchical Weighted Local Bundle Adjustment confines each BA to a limited subset of images. This strategy substantially reduces the computational cost of On-the-Fly SfM during BA, ensuring the near real-time pose estimation and sparse point cloud generation. This paper is built on the On-the-Fly SfM to update poses and sparse point cloud for our subsequent A-TDOM.

\section{Methodology}\label{Methodology}

\subsection{Overview of A-TDOM}

To generate active TDOM in near real-time, we propose a progressive online 3DGS optimization framework which enables simultaneous image acquisition and active TDOM generation. As illustrated in Fig. \ref{fig:1}, our A-TDOM can be structured into three distinct phases:

\noindent\textbf{Phase I:} In order to mitigate the risk of overfitting to the initial few images (e.g., 2 or 3 images) \citep{SparseGS} and enhance the stability of the initial 3DGS field, 3DGS optimization will wait until On-the-Fly SfM has estimated pose information and generated a sparse point cloud for $N_0$ initial images. We follow the original 3DGS optimization strategy and impose a tighter constraint on the number of training iterations to initialize 3DGS field while preventing overfitting to the initial images. This initial 3DGS field serves as the foundation for subsequent progressive training, and the initial TDOM is generated accordingly based on it.

\noindent\textbf{Phase II:} As each new image arrives, we update the current 3DGS field and the corresponding TDOM through four steps: 1) Pose registration of the incoming image and pose refinement of already registered images with On-the-Fly SfM. An updated sparse point cloud is produced. 2) 2D Delaunay triangulation on the incoming image based on reprojection of the updated sparse point cloud to determine key regions for subsequent 3DGS optimization. 3) A novel Gaussian Sampling and Integration to replace the original initialization and densification of 3DGS, appending only necessary Gaussians into the current 3DGS field. 4) 3DGS Optimization followed by TDOM generation through orthogonal splatting. 

\noindent\textbf{Phase III:} Once image capturing and progressive 3DGS optimization are complete, the quality of the 3DGS field is furtherly refined through a fast and lightweight optimization. 

\subsection{\justifying Key Region Determination via Reprojected Delaunay Triangulation Masking}

\begin{figure}[htp]
    \centering
    \includegraphics[trim={0 0 0 0},clip, width=\linewidth]{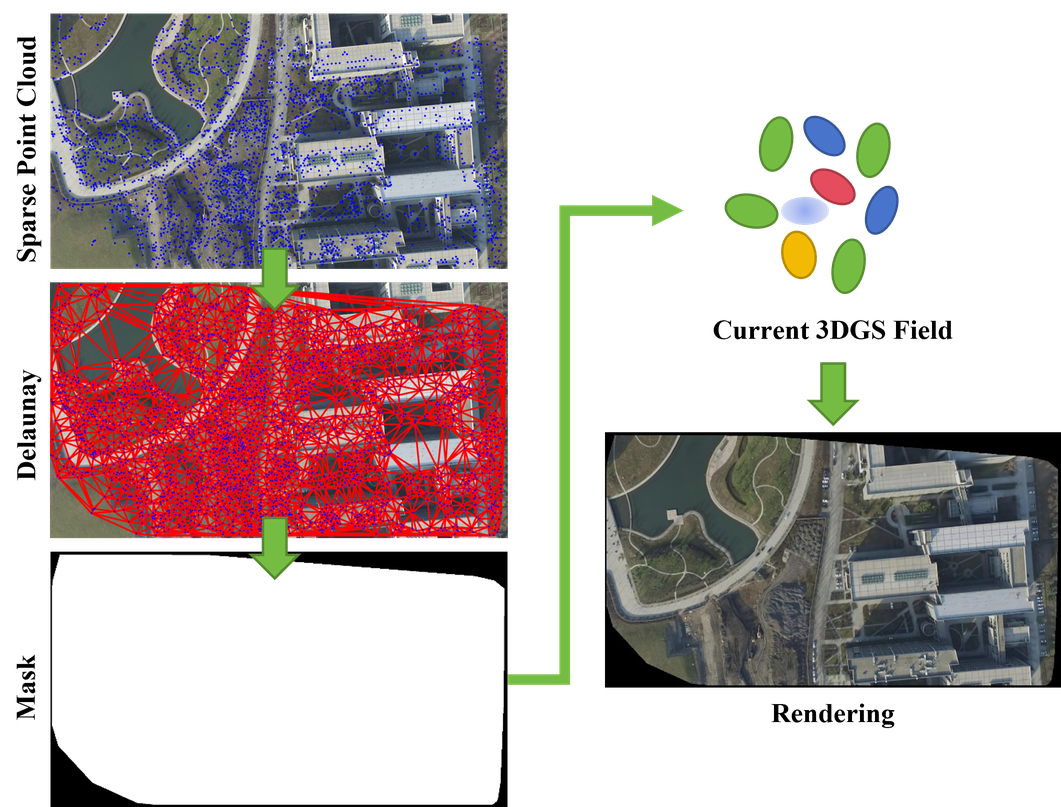}
    \caption{Scene reconstruction based on Delaunay triangular masking. A Delaunay Triangulation is constructed on the image plane based on the 3D-2D reprojections, and the regions covered by the triangulation are defined as the key regions. During subsequent training, only key regions are used for the rendering loss.}
    \label{fig:2}
    \vspace{-0.4cm}
\end{figure}

\begin{figure*}[htp]
    \centering
    \includegraphics[trim={0.65cm 0 0 0},clip, width=1.0\textwidth]{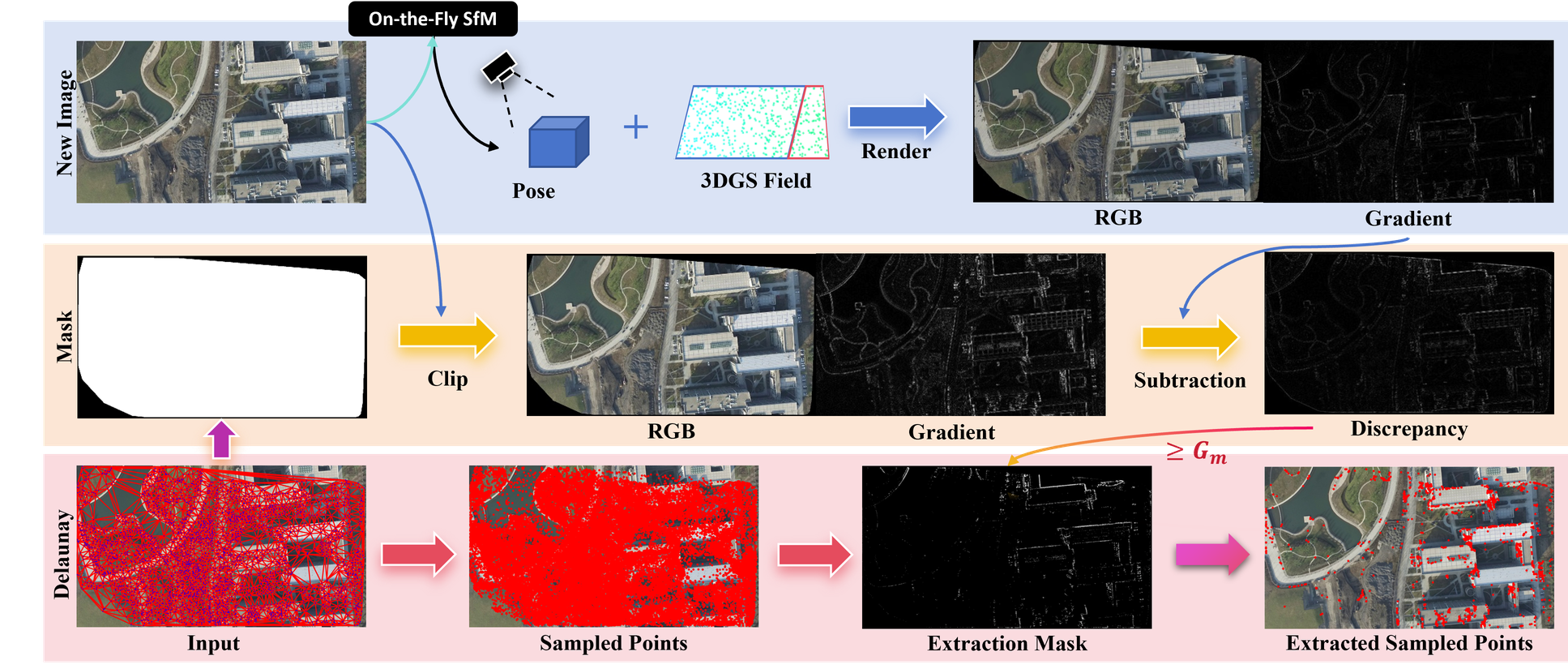}
    \caption{Gaussians Sampling and Integration. Previously unseen and coarsely reconstructed regions are identified based on gradient discrepancy map, and the extraction mask are formed. Subsequently, we sample points for Gaussians integration.}
    \label{fig:3}
\end{figure*}

Traditional TDOM methods typically consider regions with sufficient image overlap, while areas with low overlap (e.g., scene boundaries) are commonly excluded from the generated TDOM. However, the 3DGS aims to minimize the rendering loss across all images. Consequently, the trained model may attempt to reconstruct regions that are visible and overfit to only a single image or just stereo images and lack sufficient multi-view constraints. These reconstructions are suboptimal, as they tend to produce unstable or inconsistent geometry and appearance in these poorly covered areas. To address this, we estimate the key region based on the visibility between the dynamically extended sparse point cloud and each image. As illustrated in Fig. \ref{fig:2}, for each newly input image, we reproject the current sparse point cloud onto its image plane. The visible reprojections are then processed using Delaunay triangulation, which is subsequently employed to generate a validity mask. During sparse point cloud reconstruction in On-the-Fly SfM, only 3D points with high robustness are preserved, while low-quality points are eliminated during bundle adjustment \citep{On-the-Fly-SfM}. In the subsequent 3DGS optimization, only the masked image region with multi-view triangulation that can ensure sufficient overlap is optimized, while regions with insufficient overlap are excluded from training.

\subsection{\justifying Gaussians Sampling and Integration}

The original 3DGS densification strategy relies on gradient-based split-and-clone operations, where Gaussians are duplicated according to spatial gradient accumulation during training. While this approach effectively increases Gaussian density in coarsely reconstructed regions, it suffers from limited controllability and often results in excessive Gaussian accumulation that really matters in real-time applications. To overcome these limitations, instead of the original 3DGS densification strategy, we propose a novel Gaussian Sampling and Integration strategy based on the Delaunay triangulation results. Upon the acquisition of each new image, it first dynamically updates the current Gaussian field by: 1) Sampling new Gaussians in previously unseen or coarsely reconstructed regions to accelerate local optimization; 2) Avoiding redundant integration in well-reconstructed regions to reduce memory consumption and maintain computational efficiency.

As illustrated in Fig. \ref{fig:3}, for each newly input image, its camera pose and the updated sparse point cloud are computed using On-the-Fly SfM. The corresponding masked key region is then determined, within which the scene content is rendered using the current 3DGS field. Subsequently, a gradient map is computed via the Laplacian of Gaussian (LoG) operator for both the rendered image and the input clipped image. By subtracting these two gradient maps, it is possible to identify pixels exhibiting gradient discrepancies exceeding a predefined threshold $G_m$, which can be used to identify regions that requires additional Gaussians integration.

After identifying the regions $R_a$ that needs additional Gaussians integration, new Gaussians are sampled with positions and colors which are initialized based on the guidance of Delaunay triangle. Specifically, $H_t$ are uniformly sampled within each Delaunay triangle and all sampled points that fall within the region $R_a$ are retained. The retained 2D samples are then projected into 3D space. More formally, assume a sampled point $s \left ( x_s, y_s \right )$ locates in the triangle $\tau$, whose vertices’ 2D coordinates are $p_{\tau1} \left ( x_{1}, y_{1} \right )$,  $p_{\tau2} \left ( x_{2}, y_{2} \right )$ and $p_{\tau3} \left ( x_{3}, y_{3} \right )$, and the corresponding 3D coordinates are $P_{\tau1} \left ( X_{1}, Y_{1}, Z_{1} \right )$, $P_{\tau2} \left ( X_{2}, Y_{2}, Z_{2} \right )$ and $P_{\tau3} \left ( X_{3}, Y_{3}, Z_{3} \right )$. The 3D coordinate $P_{S}\left ( X_{s}, Y_{s}, Z_{s} \right )$ of the sampled point can be approximated as follows:

$$
\lambda_1 = \frac{\left ( y_{2}-y_{3} \right ) \left ( x_{s}-x_{3} \right ) + \left ( x_{3}-x_{2} \right ) \left ( y_{s}-y_{3} \right )}{ \left ( y_{2}-y_{3} \right ) \left ( x_{1}-x_{3} \right ) + \left ( x_{3}-x_{2} \right ) \left ( y_{1}-y_{3} \right )} \eqno{(4)}
$$

$$
\lambda_2 = \frac{\left ( y_{3}-y_{1} \right ) \left ( x_{s}-x_{3} \right ) + \left ( x_{1}-x_{3} \right ) \left ( y_{s}-y_{3} \right )}{ \left ( y_{2}-y_{3} \right ) \left ( x_{1}-x_{3} \right ) + \left ( x_{3}-x_{2} \right ) \left ( y_{1}-y_{3} \right )} \eqno{(5)}
$$

$$
\lambda_3 = 1 - \lambda_1 - \lambda_2 \eqno{(6)}
$$

$$
P_{S} = \lambda_1 P_{\tau1} + \lambda_2 P_{\tau2} + \lambda_3 P_{\tau3} \eqno{(7)}
$$

Similarly, assuming the colors of the triangle vertices are $C_{\tau1}$, $C_{\tau2}$ and $C_{\tau3}$, the color $C_S$ of the sampled point is initialized as follows:

$$
C_S = \lambda_1 C_{\tau1} + \lambda_2 C_{\tau2} + \lambda_3 C_{\tau3} \eqno{(8)}
$$

After estimating the 3D coordinates and colors of all sampled points, new Gaussians are initialized and integrated into the current 3DGS field, and the subsequent 3DGS optimization is performed.

\subsection{\justifying Adaptive 3DGS Optimization Strategies}

After Gaussian Sampling and Integration for each new image, a certain number of training iterations $T_I$ are performed on the current 3DGS field. The newly input images usually contain coarsely reconstructed or previously unseen regions. This implies that more training effort and higher learning rates should be applied to them. Consequently, the original 3DGS strategy for learning rate update and iteration allocation becomes infeasible in such a dynamic scenario. Inspired by our previous Gaussian On-the-Fly Splatting \citep{On-the-Fly-GS}, several modifications are applied to the standard 3DGS training. \textbf{1) Iteration allocation:} Since the most recently input image typically contains the largest proportion of previously unseen regions, half of the training iterations $\frac{T_I }{2}$ are assigned to this image during the optimization of the 3DGS field. Meanwhile, to ensure continuous optimization of the other registered images, the remaining $\frac{T_I}{2}$ training iterations are evenly distributed among these registered images; \textbf{2) Learning rate update:} Since the regions corresponding to the most recent input and to the previously registered images are optimized to different extents, learning rates are assigned per image and the learning rate decay is handled independently for each image. For images with high rendering quality, a lower learning rate is assigned to ensure stable refinement since their corresponding local 3DGS fields have already been well optimized. In contrast, images with poor rendering quality are trained with a higher learning rate to promote faster convergence. \textbf{3) Final Refinement.} After all images have been acquired and trained, we perform a rapid refinement on the current 3DGS field, as illustrated in Phase III of Fig. \ref{fig:1}. During this stage, the training iterations are evenly distributed across all images to ensure that every region of the 3DGS field receives further optimization. The learning rate is still assigned per image which remains consistent with Phase II, allowing images with lower rendering quality to be optimized more aggressively. Finally, the Gaussian densification strategy adopted in the original 3DGS training is still disabled, ensuring that the existing Gaussians are optimized more effectively without introducing redundant primitives.

\subsection{\justifying Orthographic Splatting for Active TDOM Generation}

The vanilla 3DGS method employs perspective transformation to project 3D Gaussians onto 2D image plane for rendering, which inevitably compresses and distorts the 3D scene geometry. In contrast, orthographic transformation preserves the true geometric proportions of ground objects in both size and shape, offering substantial potential for generating correctly rectified TDOM. Building upon the Gaussian splatting method \citep{zwicker2001ewa}, we employ orthographic splatting that incorporates the orthographic projection of both the mean and covariance of Gaussians.

\begin{figure*}[ht]
    \centering
    \includegraphics[width=0.8\textwidth]{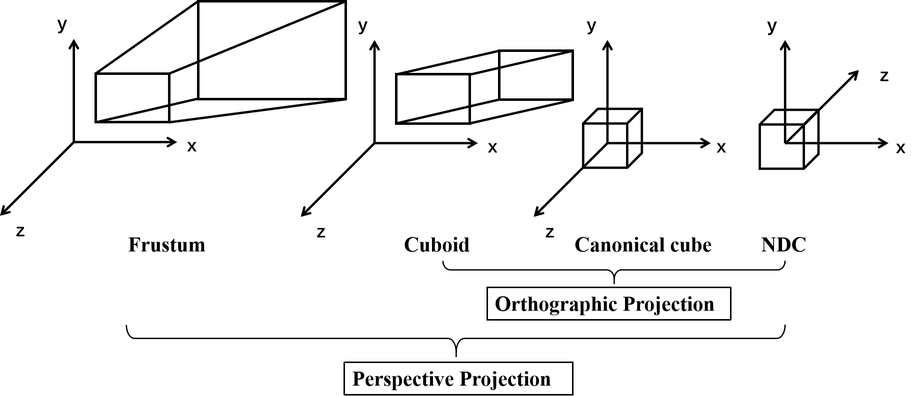}
    \caption{Projection Transformation Process. In perspective projection, the Gaussians are projected from the frustum, whereas in orthographic projection, they are projected from the cuboid, both into the Normalized Device Coordinates (NDC) system.}
    \label{fig:projection}
\end{figure*}

\textbf{Projection of mean:}  The perspective projection matrix \textit{P} consists of three sequential transformations as illustrated in Fig. \ref{fig:projection}: the Frustum-to-Cuboid Transformation($M_{persp2ortho}$), the Cuboid-to-Canonical Cube Transformation($M_{ortho}$) and Canonical Cube-to-NDC Transformation($M_{hand}$). 
It can be formulated as:
$$
\begin{aligned}
    P &=M_{hand}M_{ortho}M_{persp2ortho}\\
    &=\begin{pmatrix}
      \frac{2n}{r-l}&0&-\frac{r+l}{r-l}&0\\0&\frac{2n}{t-b}&-\frac{t+b}{t-b}&0\\0&0&-\frac{n+f}{n-f}&\frac{2nf}{n-f}\\0&0&1&0 
  \end{pmatrix} 
\end{aligned}\eqno{(9)}
$$
where
$$
    r=tan(\frac{FovX}{2})*n, t=tan(\frac{FovY}{2})*n \eqno{(10)}
$$

$n$, $f$, $l$, $r$, $t$, $b$ represent the near, far, left, right, top, bottom boundaries of the viewing frustum, while $FovX$ and $FovY$ denote the horizontal and vertical fields of view, respectively. 

To perform orthographic projection, the term $M_{persp2ortho}$ is excluded from the above composition. The resulting orthographic matrix $P_0$ is expressed as:
$$
    P_0
    =\begin{pmatrix}
        \frac{2}{r-l}&0&0&-\frac{r+l}{r-l}\\0&\frac{2}{t-b}&0&-\frac{t+b}{t-b}\\0&0&-\frac{2}{n-f}&\frac{n+f}{n-f}\\0&0&0&1
    \end{pmatrix} \eqno{(11)}
$$
Via this formulation, the center of Gaussians are orthographically projected onto 2D image plane.

\textbf{Projection of Covariance:}  In the perspective projection, the $M_{persp2ortho}$ matrix is non-affine, requiring an affine approximation of the Gaussian covariance transformation. The covariance matrix $\Sigma$, is therefore transformed using the Jacobian $J$ derived from equation (9):
$$
    J=\begin{pmatrix}
        \frac{focal_x}{z}&0&-\frac{focal_xx}{z^2}\\0&\frac{focal_y}{z}&-\frac{focal_yy}{z^2}\\0&0&0
    \end{pmatrix}\eqno{(12)}
$$
where $\left (x, y, z\right )$ denotes the 3D coordinates of a point in the camera coordinate system, and $focal_x$, $focal_y$ are the focal lengths along the X and Y axes, respectively.

Since orthographic projection is an affine transformation, its Jacobian can be directly given by the equation(11):
$$
J_0=
    \begin{pmatrix}
        \frac{2}{r-l}&0&0\\0&\frac{2}{t-b}&0\\0&0&0
    \end{pmatrix}\eqno{(13)}
$$

\textbf{$\alpha$-blending using orthogonal projection:} By employing $P_0$ and $J_0$, the Gaussian mean and covariance can be accurately projected onto the NDC coordinate system, where visibility clipping is subsequently performed. The Gaussians are then transformed into the pixel space via the viewport transformation, where the Gaussians that intersect the corresponding GSD (Ground Sampling Distance) are considered and their transparency values are computed according to equation (2), and the final pixel color is obtained by $\alpha$-blending as defined in equation (3).

\textbf{GSD:}
The rendering resolution was determined by the Ground Sample Distance (GSD). specifically, given the scene's bounding box dimensions ($X_{max}, X_{min}, Y_{max}, Y_{min}$) and the desired GSD, the output image dimensions ($W, H$) was calculated as:
$$
W=\frac{X_{max}-X_{min}}{GSD_x} \eqno{(14)}
$$
$$
H=\frac{Y_{max}-Y_{min}}{GSD_y} \eqno{(15)}
$$
In the orthogonal rendering pipeline, the ground point coordinates sequentially undergo $World-to-View$, $Orthographic Projection$, and the $Viewport$ transformations. Based on the cumulative scaling relationships among these transformations, the key parameter can be derived:
$$
FovX=2\arctan(\frac{GSD_x*W*scale}{-2f*ratio})\eqno{(16)}
$$
$$
FovY=2\arctan(\frac{GSD_y*H*scale}{-2f})\eqno{(17)}
$$
where \textit{scale} denotes a fixed scaling factor in the world coordinate system and $ratio$ represents the aspect ratio of individual pixels.

\section{Experiment}\label{Experiments}

In our A-TDOM, a novel near real-time 3DGS training method is proposed, built on which TDOM can be actively rendered in near real time. We evaluate these two components separately to highlight the advantages of A-TDOM.

\subsection{Experimental Settings}

\textbf{Near real-time 3DGS training.} To demonstrate A-TDOM's perfomance and adaptability across diverse scenarios in near real-time 3DGS training, we conduct several tests on Replica datasets (RGB-D SLAM) \citep{Replica}, small-scale scene datasets for Novel View Synthesis (NVS) and UAV image datasets. Replica provides image sequences with strong spatiotemporal continuity, making it well suited for evaluating SLAM-based methods. Eight scenes from Replica are selected for our subsequent evaluation. For small-scale scene datasets, we test on nine indoor and outdoor scenes provided by Mip-NeRF 360 (RGB datasets) \citep{Mip-NeRF-360} and On-the-Fly SfM \citep{On-the-Fly-SfM}. For UAV image datasets, we evaluate A-TDOM on nine selected scenes from BeDOI \citep{BeDOI}, CAS \citep{NPU_AND_PHANTOM3} and On-the-Fly SfM. Based on these datasets, A-TDOM is compared against two classical SLAM-based 3DGS real-time training solutions (Gaussian Splatting SLAM \citep{GS-SLAM} and RTG-SLAM \citep{RTG-SLAM}), one notable efficient offline training solution (AdR-Gaussian \citep{AdR-Gaussian}, trained for \textbf{7K iterations} due to the efficiency-oriented focus of this evaluation), On-the-Fly NVS \citep{On-the-Fly-NVS} and Gaussian On-the-Fly Splatting \citep{On-the-Fly-GS}. The corresponding performance of each method is evaluated via three aspects: 1) Rendering quality. The metrics commonly used for 3DGS-based solutions evaluation, including PSNR, SSIM and LPIPS, are adopted. 2) Training efficiency. The FPS of different real-time/near real-time 3DGS training solutions is recorded to reveal the time efficiency. 3) Model size. The size of each 3DGS model generated by the different solution is compared to evaluate memory consumption and model complexity. All experiments are conducted under the same hardware configuration (Single NVIDIA RTX 4090 GPU 24G).

\begin{figure*}[ht]
    \centering
    \includegraphics[width=\textwidth]{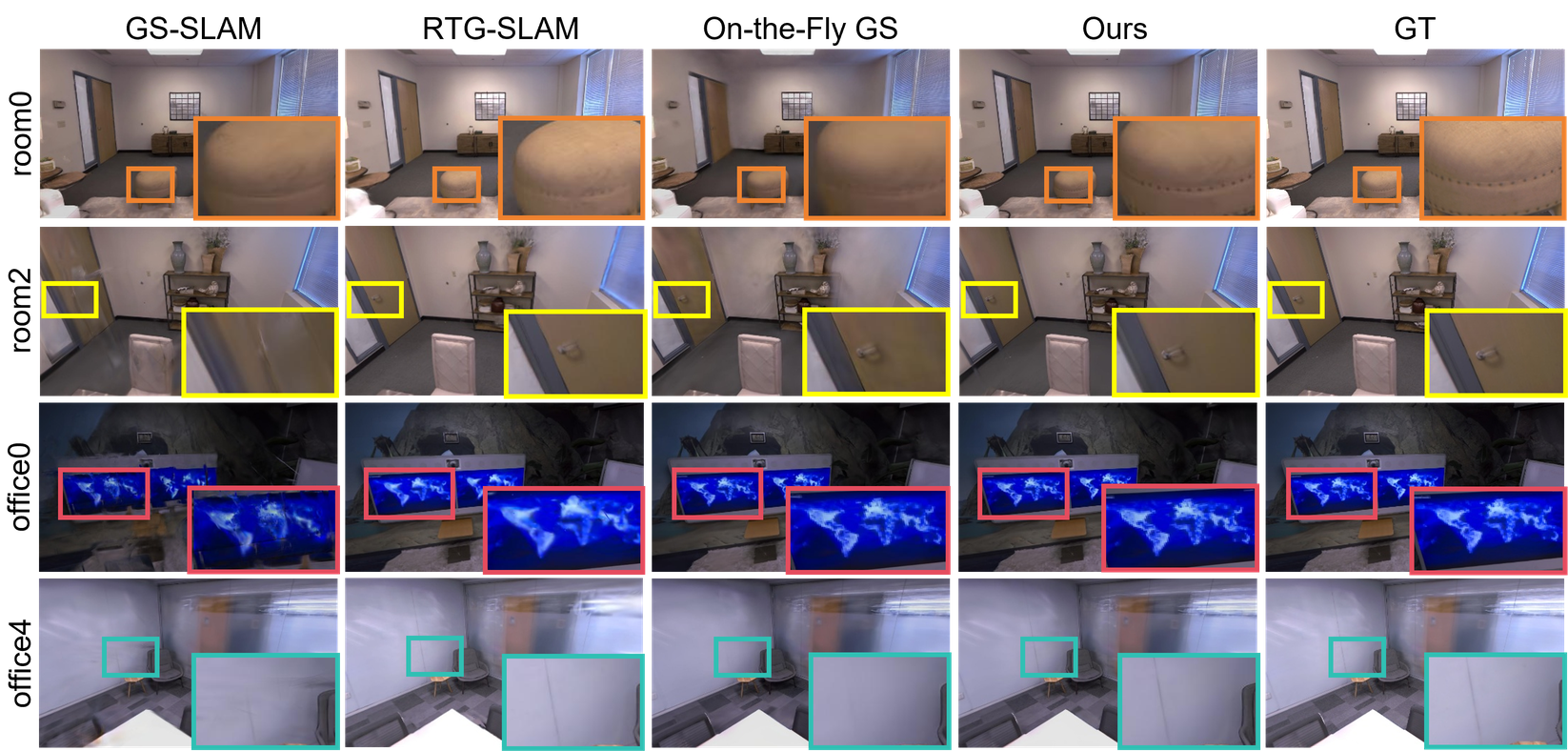}
    \caption{Rendering Results of A-TDOM and other solutions with default settings on SLAM datasets. Our A-TDOM achieves significantly higher rendering quality than other solutions.}
    \label{Replica_Fig}
\end{figure*}

\textbf{TDOM generation.} The application goal of this paper is to enable near real-time TDOM generation. Therefore, in addition to evaluating A-TDOM from the perspective of NVS, we further assess its performance in TDOM generation. We compare our A-TDOM with three offline 3DGS-based methods (AdR-Gaussian \citep{AdR-Gaussian}, Tortho-Gaussian \citep{Tortho-gaussian} and VastGaussian \citep{VastGaussian}) and three commercial software (ContextCapture, MetaShape and Pix4DMapper) on part of UAV image datasets. Since this evaluation focuses on TDOM generation, to clearly reflect the trade-off between rendering quality and computational efficiency, AdR-Gaussian, Tortho-Gaussian, and VastGaussian are all trained for \textbf{30K iterations} in the corresponding experiment. The subsequent experiments are conducted both qualitatively and quantitatively. In the qualitative evaluation, we mainly focus on the overall visual quality of the generated TDOMs and the effectiveness of facade removal, particularly for building structures. In the quantitative evaluation, we first compare the time required for TDOM generation across different methods. For the 3DGS-based methods, we further analyze their performance in terms of final rendering quality and model size. In addition, since the \textit{npu} scene provides ground-truth control points, we also evaluate the geometric accuracy of the generated TDOM using different methods.

\textbf{Implementation details.} As illustrated in Fig. \ref{fig:1}, after On-the-Fly SfM estimates the pose information and generates the sparse point cloud using the initial $N_0 = 30$ images, A-TDOM performs 2000 training iterations on the initial 3DGS field and produces the initial TDOM. In phase II, for each newly input image, its camera pose and the extended sparse point cloud are estimated via On-the-Fly SfM, followed by the Delaunay triangulation on the newly input image. During Gaussian Sampling and Integration, the pixels exhibiting gradient discrepancies exceeding $G_m=0.1$ are identified as the regions that requires additional Gaussians integration. When sampling new Gaussians, the number of sampling points per Delaunay triangle, denoted as $H_t$, is assigned based on scene complexity. In this paper, we set $H_t = 10$ for all scenes in the Replica dataset, $H_t = 10$ or $20$ for the small-scale scene datasets, and $H_t = 20$ or $30$ for the UAV image datasets. After Gaussian Sampling and Integration, a very fast optimization of the current 3DGS field for $T_I = 200$ iterations is performed and the TDOM is updated immediately.

Additionally, since both A-TDOM and Gaussian On-the-Fly Splatting rely on pose information and sparse point cloud estimated by On-the-Fly SfM for 3DGS training, for a fair comparison, all other offline 3DGS-based methods that require pre-estimated pose information and sparse point cloud are also provided by On-the-Fly SfM. In the TDOM generation experiments, as VastGaussian adopts a scene partitioning strategy that enables parallel training, we fix the number of partitions to 4 for all scenes in the subsequent evaluations.

\subsection{\justifying The Performance of Real-Time 3DGS Training}

In this section, based on the SLAM dataset, small-scale indoor or outdoor scene datasets, and UAV image datasets, we conduct comparative experiments between our A-TDOM and various 3DGS methods including both online SLAM-based and offline 3DGS methods. 

\begin{table*}[ht]
\centering
\caption{\justifying Comparison on Replica SLAM datasets. $\uparrow$ indicates higher is better, while $\downarrow$ indicates lower is better. Best and second-best are highlighted in \textbf{bold} and \underline{underlined}.}
\label{Replica_Tab}
\resizebox{\textwidth}{!}{
\begin{tabular}{cccccccccccc}

\toprule

Method                         & Modality               & Metric          & room0           & room1           & room2           & office0         & office1         & office2         & office3         & office4         & Average         \\ \midrule
\multirow{6}{*}{GS-SLAM}       & \multirow{6}{*}{RGB}   & PSNR (dB)$\uparrow$& 29.31           & 27.58           & 29.24           & 33.88           & 35.51           & 25.24           & 28.79           & 27.98           & 29.89           \\
                               &                        & SSIM$\uparrow$  & 0.89            & 0.84            & 0.89            & 0.93            & 0.94            & 0.88            & 0.89            & 0.91            & 0.89            \\
                               &                        & LPIPS$\downarrow$& 0.24            & 0.32            & 0.23            & 0.18            & 0.16            & 0.28            & 0.18            & 0.24            & \underline{0.22}      \\
                               &                        & Model Size (MB)$\downarrow$& \textbf{8.9}    & \textbf{7.3}    & \textbf{4.6}    & \textbf{4.5}    & \textbf{3.9}    & \textbf{6.4}    & \textbf{6.9}    & \textbf{5.2}    & \textbf{5.9}    \\
                               &                        & FPS$\uparrow$   & 0.89            & 0.92            & 0.93            & 0.95            & 1.05            & 0.89            & 0.93            & 0.91            & 0.93            \\
                               &                        & ATE-MSE$\downarrow$& 0.0849          & 0.4468          & 0.1897          & 0.2014          & 0.2144          & 0.5454          & 0.1664          & 0.8466          & 0.1869          \\ \midrule
\multirow{6}{*}{RTG-SLAM}      & \multirow{6}{*}{RGB-D} & PSNR (dB)$\uparrow$& 31.56           & \underline{34.21}     & \underline{35.57}     & 39.11           & \textbf{40.27}  & \underline{33.54}     & 32.67           & 36.48           & 35.43           \\
                               &                        & SSIM$\uparrow$  & \underline{0.97}      & \textbf{0.98}   & \underline{0.98}      & \textbf{0.99}   & \textbf{0.99}   & \textbf{0.98}   & \textbf{0.98}   & \underline{0.98}      & \underline{0.98}      \\
                               &                        & LPIPS$\downarrow$& \underline{0.13}      & \textbf{0.11}   & \underline{0.12}      & \textbf{0.07}   & \underline{0.08}      & \underline{0.13}      & 0.13            & 0.12            & \textbf{0.11}   \\
                               &                        & Model Size (MB)$\downarrow$& 53.1            & 72.6            & 57.4            & 46.7            & 53.1            & 49.2            & 51.6            & 54.7            & 54.8            \\
                               &                        & FPS$\uparrow$   & \underline{2.71}      & \underline{2.94}      & \underline{2.73}      & 2.71            & 3.05            & \underline{3.13}      & \underline{2.94}      & \underline{2.94}      & \underline{2.89}      \\
                               &                        & ATE-MSE$\downarrow$& \textbf{0.0019} & \textbf{0.0019} & \textbf{0.0011} & \textbf{0.0015} & \textbf{0.0013} & \textbf{0.0024} & \textbf{0.0023} & \textbf{0.0025} & \textbf{0.0019} \\ \midrule
\multirow{6}{*}{On-the-Fly GS} & \multirow{6}{*}{RGB}   & PSNR (dB)$\uparrow$& \underline{31.79}     & 33.67           & 34.65           & \underline{39.44}     & 38.77           & \textbf{34.03}  & \textbf{35.38}  & \underline{36.65}     & \underline{35.55}     \\
                               &                        & SSIM$\uparrow$   & 0.94            & \underline{0.95}      & 0.96            & \underline{0.97}      & \underline{0.96}      & \underline{0.95}      & \underline{0.96}      & 0.96            & 0.96            \\
                               &                        & LPIPS$\downarrow$& \textbf{0.12}   & \textbf{0.11}   & \textbf{0.11}   & \underline{0.09}      & 0.14            & \underline{0.13}      & \textbf{0.09}   & \underline{0.10}      & \textbf{0.11}   \\
                               &                        & Model Size (MB)$\downarrow$& 130.1           & 111.7           & 69.4            & 36.7            & 28.6            & 69.7            & 62.1            & 56.5            & 70.6            \\
                               &                        & FPS$\uparrow$    & 2.01            & 2.23            & 1.99            & \underline{3.25}      & \underline{3.56}      & 1.86            & 1.88            & 2.12            & 2.36            \\
                               &                        & ATE-MSE$\downarrow$& \underline{0.0035}    & \underline{0.0054}    & \underline{0.0025}    & \underline{0.0057}    & \underline{0.0053}    & \underline{0.0070}    & \underline{0.0024}    & \underline{0.0075}    & \underline{0.0049}    \\ \midrule
\multirow{6}{*}{Ours}          & \multirow{6}{*}{RGB}   & PSNR (dB)$\uparrow$& \textbf{33.65}  & \textbf{34.22}  & \textbf{35.82}  & \textbf{41.09}  & \underline{39.65}     & 33.33           & \underline{33.84}     & \textbf{37.31}  & \textbf{36.11}  \\
                               &                        & SSIM$\uparrow$  & \textbf{0.98}   & \textbf{0.98}   & \textbf{0.99}   & \textbf{0.99}   & \textbf{0.99}   & \textbf{0.98}   & \textbf{0.98}   & \textbf{0.99}   & \textbf{0.99}   \\
                               &                        & LPIPS$\downarrow$& 0.14            & \underline{0.13}      & 0.13            & \textbf{0.07}   & \textbf{0.07}   & \textbf{0.10}   & \underline{0.11}      & \textbf{0.09}   & \textbf{0.11}   \\
                               &                        & Model Size (MB)$\downarrow$& \underline{30.1}      & \underline{45.6}      & \underline{24.1}      & \underline{34.8}      & \underline{25.9}      & \underline{22.2}      & \underline{18.1}      & \underline{26.7}      & \underline{28.4}      \\
                               &                        & FPS$\uparrow$   & \textbf{5.08}   & \textbf{5.41}   & \textbf{4.53}   & \textbf{4.66}   & \textbf{6.46}   & \textbf{4.65}   & \textbf{4.34}   & \textbf{4.84}   & \textbf{4.99}   \\
                               &                        & ATE-MSE$\downarrow$& \underline{0.0035}    & \underline{0.0054}    & \underline{0.0025}    & \underline{0.0057}    & \underline{0.0053}    & \underline{0.0070}    & \underline{0.0024}    & \underline{0.0075}    & \underline{0.0049}    \\          

\bottomrule

\end{tabular}
}
\end{table*}


\textbf{Experiments on SLAM datasets.} Since the Replica dataset provides image sequences with extremely high overlap of strong spatiotemporal continuity, Gaussian Splatting SLAM (abbreviated as GS-SLAM) and RTG-SLAM extract keyframes which are just used for 3DGS optimization. Therefore, the keyframes selected by GS-SLAM are preserved and reused as input for both Gaussian On-the-Fly Splatting (abbreviated as On-the-Fly GS) and our A-TDOM. Moreover, the Replica dataset is with ground-truth pose information, we further evaluate the pose estimation accuracy of different methods by computing the Mean Squared Error of the Absolute Trajectory Error (ATE-MSE). 

As illustrated in Fig. \ref{Replica_Fig} and summarized in Tab. \ref{Replica_Tab}, compared to other methods, A-TDOM achieves the best or second-best rendering quality in most scenes while also delivering the highest training efficiency. In terms of tracking accuracy, both A-TDOM and our previous On-the-Fly GS leverage the On-the-Fly SfM for pose estimation, thereby exhibiting identical tracking performance, which is superior than GS-SLAM and inferior to RTG-SLAM that takes depth prior as input for better tracking.

\begin{figure*}[ht]
    \centering
    \includegraphics[width=\textwidth]{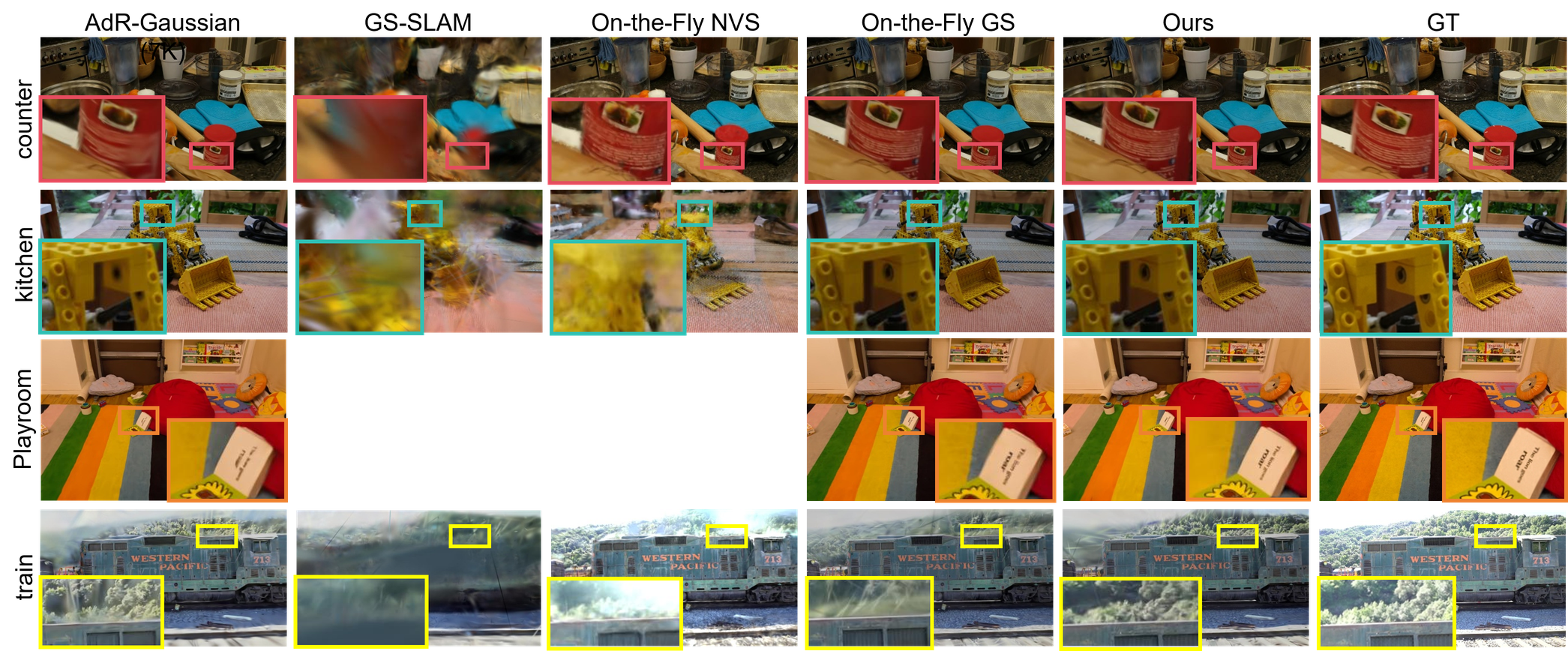}
    \caption{Rendering Results of A-TDOM and other solutions with default settings on small-scale scene datasets. Our A-TDOM achieves significantly higher rendering quality than other solutions. Blank indicates the tracking failure.}
    \label{small_scale_scene _Fig}
\end{figure*}

\begin{table*}[ht]
\centering
\caption{\justifying Comparison on small scale scene datasets. $\uparrow$ indicates higher is better, while $\downarrow$ indicates lower is better. Best and second-best are highlighted in \textbf{bold} and \underline{underlined}. $\setminus$ denotes tracking failure due to low spatiotemporal continuity.}
\label{small_scale_scene_Tab}
\resizebox{\textwidth}{!}{
\begin{tabular}{cc|ccccccccc|c}

\toprule

\multicolumn{2}{c|}{Scene}                           & counter        & kitchen        & stump                     & treehill                  & train          & Caffe          & Playroom                          & JYYL           & SRSX           & \multirow{3}{*}{Avg.} \\
\multicolumn{2}{c|}{Image Number}                    & 240            & 279            & 125                       & 141                       & 301            & 159            & 222                               & 390            & 221            &                       \\
\multicolumn{2}{c|}{Pose Estimation (s)}             & 387.09         & 606.52         & 173.61                    & 216.92                    & 412.33         & 324.49         & 238.71                            & 735.85         & 394.64         &                       \\ \midrule
\multirow{5}{*}{AdR-Gaussian (7K)} & PSNR (dB)$\uparrow$& 26.81          & 28.45          & 24.31                     & \underline{20.31}               & 19.94          & \underline{18.81}    & 32.09                             & 19.85          & 24.71          & 23.92                 \\
                                   & SSIM$\uparrow$  & 0.86           & 0.89           & 0.89                      & \underline{0.81}                & 0.69           & 0.76           & \textbf{0.99}                     & 0.69           & \underline{0.82}     & 0.82                  \\
                                   & LPIPS$\downarrow$& 0.24           & \underline{0.15}     & 0.33                      & \underline{0.42}                & 0.36           & 0.39           & 0.31                              & 0.47           & \underline{0.24}     & 0.32                  \\
                                   & Model Size (MB)$\downarrow$& 104.5          & 192.5          & \underline{321.1}               & \underline{162.1}               & \underline{63.2}     & \underline{91.5}     & 178.1                             & \underline{92.8}     & 320.2          & 169.6                 \\
                                   & Cost Time (s)$\downarrow$& 595.05         & 852.48         & 317.15                    & 353.35                    & 521.69         & 516.55         & 390.1                             & 1016.43        & 648.47         & 579.03                \\ \midrule
\multirow{5}{*}{GS-SLAM}           & PSNR (dB)$\uparrow$& 18.19          & 17.59          & \multicolumn{2}{c}{\multirow{5}{*}{\textbackslash{}}} & 16.34          & 15.03          & \multirow{5}{*}{\textbackslash{}} & 18.25          & 19.11          & 17.42                 \\
                                   & SSIM$\uparrow$  & 0.63           & 0.48           & \multicolumn{2}{c}{}                                  & 0.49           & 0.43           &                                   & 0.66           & 0.39           & 0.51                  \\
                                   & LPIPS$\downarrow$& 0.74           & 0.76           & \multicolumn{2}{c}{}                                  & 0.72           & 0.79           &                                   & 0.62           & 0.94           & 0.76                  \\
                                   & Model Size (MB)$\downarrow$& \textbf{22.6}  & \textbf{43.6}  & \multicolumn{2}{c}{}                                  & \textbf{23.7}  & \textbf{7.6}   &                                   & \textbf{26.1}  & \textbf{4.4}   & \textbf{21.3}         \\
                                   & FPS$\uparrow$   & 0.19           & 0.19           & \multicolumn{2}{c}{}                                  & 0.37           & 0.26           &                                   & 0.26           & 0.28           & 0.26                  \\ \midrule
\multirow{5}{*}{On-the-Fly NVS}    & PSNR (dB)$\uparrow$& 27.01          & 17.94          & \multicolumn{2}{c}{\multirow{5}{*}{\textbackslash{}}} & 18.67          & 18.12          & \multirow{5}{*}{\textbackslash{}} & 20.79          & \underline{26.09}    & 21.44                 \\
                                   & SSIM$\uparrow$     & 0.88           & 0.63           & \multicolumn{2}{c}{}                                  & 0.64           & 0.58           &                                   & 0.74           & \underline{0.82}     & 0.72                  \\
                                   & LPIPS$\downarrow$& 0.21           & 0.43           & \multicolumn{2}{c}{}                                  & 0.37           & 0.39           &                                   & 0.35           & 0.25           & 0.33                  \\
                                   & Model Size (MB)$\downarrow$& 143.3          & 117.7          & \multicolumn{2}{c}{}                                  & 509.4          & 140.9          &                                   & 486.3          & 237.5          & 272.5                 \\
                                   & FPS$\uparrow$   & \textbf{2.38}  & \textbf{2.93}  & \multicolumn{2}{c}{}                                  & \textbf{2.76}  & \textbf{2.91}  &                                   & \textbf{2.55}  & \textbf{2.26}  & \textbf{2.63}         \\ \midrule
\multirow{5}{*}{On-the-Fly GS}     & PSNR (dB)$\uparrow$& \textbf{28.97} & \textbf{30.39} & \underline{24.79}               & 20.24                     & \underline{22.82}    & \textbf{20.18} & \textbf{34.27}                    & \textbf{22.71} & 25.17          & \underline{25.51}           \\
                                   & SSIM$\uparrow$     & \underline{0.89}     & \underline{0.91}     & \underline{0.91}                & 0.79                      & \underline{0.81}     & \underline{0.79}     & \textbf{0.99}                     & \underline{0.79}     & 0.79           & \underline{0.85}            \\
                                   & LPIPS$\downarrow$& \underline{0.19}     & \textbf{0.11}  & \underline{0.29}                & 0.45                      & \underline{0.21}     & \underline{0.32}     & \underline{0.27}                        & \underline{0.28}     & 0.25           & \underline{0.26}            \\
                                   & Model Size (MB)$\downarrow$& 46.45          & 176.4          & 358.3                     & 207.9                     & 137.4          & 254.3          & \underline{167.3}                       & 416.4          & 367.4          & 236.9                 \\
                                   & FPS$\uparrow$   & 0.39           & 0.27           & \underline{0.52}                      & \underline{0.52}                      & 0.37           & 0.34           & \underline{0.49}                              & 0.28           & 0.31           & 0.39                  \\ \midrule
\multirow{5}{*}{Ours}              & PSNR (dB)$\uparrow$& \underline{28.09}    & \underline{30.31}    & \textbf{25.88}            & \textbf{21.28}            & \textbf{23.91} & \textbf{20.18} & \underline{34.15}                       & \underline{22.31}    & \textbf{27.34} & \textbf{25.94}        \\
                                   & SSIM$\uparrow$     & \textbf{0.96}  & \textbf{0.96}  & \textbf{0.95}             & \textbf{0.89}             & \textbf{0.91}  & \textbf{0.84}  & \textbf{0.99}                     & \textbf{0.91}  & \textbf{0.83}  & \textbf{0.92}         \\
                                   & LPIPS$\downarrow$& \textbf{0.18}  & \underline{0.15}     & \textbf{0.17}             & \textbf{0.22}             & \textbf{0.15}  & \textbf{0.22}  & \textbf{0.14}                     & \textbf{0.19}  & \textbf{0.23}  & \textbf{0.18}         \\
                                   & Model Size (MB)$\downarrow$& \underline{38.1}     & \underline{117.3}    & \textbf{154.5}            & \textbf{161.8}            & 115.5          & 198.8          & \textbf{81.3}                     & 131.9          & \underline{67.9}     & \underline{118.6}           \\
                                   & FPS$\uparrow$      & \underline{0.62}     & \underline{0.46}     & \textbf{0.72}             & \textbf{0.65}             & \underline{0.73}     & \underline{0.49}     & \textbf{0.93}                     & \underline{0.53}     & \underline{0.56}     & \underline{0.63}           \\

\bottomrule

\end{tabular}
}
\end{table*}

Among SLAM-based methods, GS-SLAM supports both RGB and RGB-D input, whereas RTG-SLAM only accepts RGB-D input. In the RGB-only setting, A-TDOM consistently achieves significantly superior rendering quality, tracking accuracy, and training efficiency than GS-SLAM in all scenes. Although RTG-SLAM achieves the highest tracking accuracy, A-TDOM obtains better rendering quality, smaller model size, and higher training efficiency without prior depth information. These results demonstrate that even on SLAM datasets, A-TDOM can outperform SLAM-based methods without requiring depth measurements.

\begin{figure*}[ht]
    \centering
    \includegraphics[width=\textwidth]{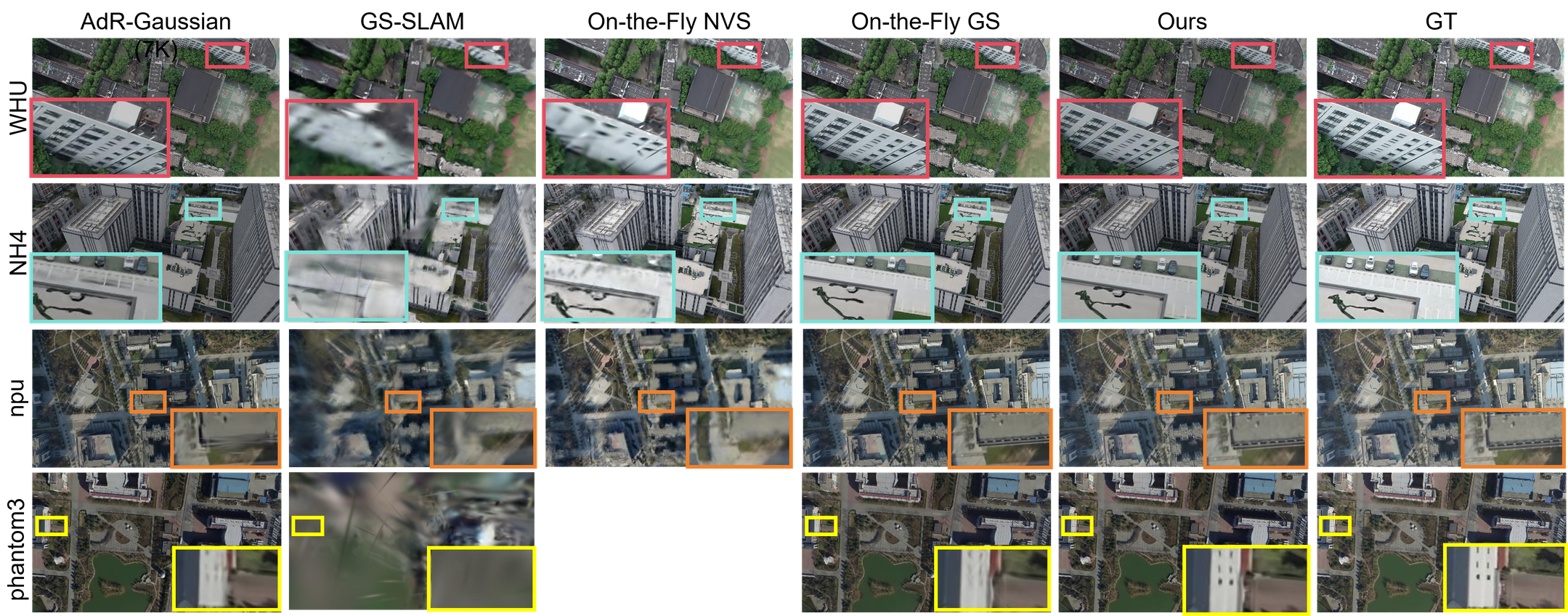}
    \caption{Rendering Results of A-TDOM and other solutions with default settings on UAV image datasets. Compared with AdR-Gaussian, GS-SLAM, and On-the-Fly NVS, our A-TDOM achieves substantially higher rendering quality. Blank indicates training failure due to out of memory.}
    \label{UAV_images_Fig}
\end{figure*}

\textbf{Experiments on small-scale scene datasets.} Compared to the SLAM datasets, the small-scale scene datasets provide image collections with lower image overlaps of less spatiotemporal continuity. Therefore, each image is employed as keyframe for both GS-SLAM and On-the-Fly NVS. Meanwhile, A-TDOM and On-the-Fly GS also perform pose estimation, sparse point cloud generation, and near real-time 3DGS training on all images. For the offline training solution, for a fair comparison, AdR-Gaussian leverages the SfM results obtained from the On-the-Fly SfM after processing all images. The total runtime for AdR-Gaussian is measured as the sum of the pose estimation time of On-the-Fly SfM and the subsequent 3DGS training time. Notably, as the small-scale scene and UAV image datasets only provide RGB input, RTG-SLAM is excluded from the following experiments.

Tab. \ref{small_scale_scene_Tab} and Fig. \ref{small_scale_scene _Fig} summarize the results quantitatively and qualitatively, respectively. Compared to the AdR-Gaussian with 7K iterations, A-TDOM shows obvious advantages in rendering quality, model size, and training efficiency. In addition, compared to our previous On-the-Fly GS, A-TDOM significantly reduces model size and improves training efficiency while maintaining comparable or slightly better rendering quality. These results demonstrate that the effectiveness of the proposed Gaussian Sampling and Integration strategy together with the Adaptive 3DGS Optimization Strategies, enabling more precise Gaussian insertion and more efficient optimization. Therefore, the number of Gaussians can be minimized without sacrificing rendering quality, leading to less memory consumption and superior training efficiency. 

Due to the fact that the images from the \textit{stump}, \textit{treehill}, and \textit{Playroom} are unordered, GS-SLAM and On-the-Fly NVS just fail during the tracking. This indicates that both SLAM-based methods and On-the-Fly NVS are highly dependent on image sequences with high spatiotemporal continuity, which could limit their real applicability in certain scenarios. For the other scenes consisting of ordered image sequences, the limited spatiotemporal continuity, lower overlap, and the absence of prior depth information lead to extremely poor rendering quality for GS-SLAM across all scenes. Furthermore, the higher image resolution and more complex details of the selected small-scale scene datasets significantly degrade the training efficiency of GS-SLAM when comparing to its corresponding performance on SLAM datasets. These experimental results reveals that SLAM-based methods perform well only when processing on SLAM datasets with very high overlapping and struggle to obtain satisfactory performance on other general datasets.

On-the-Fly NVS achieves the highest FPS among all methods, demonstrating a clear advantage in runtime efficiency when processing ordered image sequences. However, compared to A-TDOM, On-the-Fly NVS consistently exhibits slightly inferior rendering quality and significantly larger model size across all scenes. In addition, the noticeably poor performance on the \textit{kitchen} scene indicates potential robustness and stability issues under certain conditions. In contrast, A-TDOM maintains stable rendering quality across all scenes, highlighting its advantages over On-the-Fly NVS.

\textbf{Experiments on UAV image datasets.} For the evaluations on the UAV image datasets, we follow the same experimental protocol as the small-scale scene datasets. The experimental results are presented in Tab. \ref{UAV_images_Tab} and Fig. \ref{UAV_images_Fig}. It can be found that Our A-TDOM achieves the best overall performance on most scenes. More specifically, for GS-SLAM and On-the-Fly NVS, no tracking failures occur since all UAV image datasets provide ordered image sequences with a certain degree of overlapping. However, due to the high image resolution and the complex texture of UAV images, GS-SLAM exhibits both low rendering quality and inferior training efficiency. Although On-the-Fly NVS performs well on the \textit{NH2} scene, it still exhibits inferior rendering quality on the other scenes, while GPU memory constraints even prevent it completing the 3DGS training on the \textit{phantom3}, \textit{ground}, and \textit{park} scenes.

\begin{table*}[ht]
\centering
\caption{\justifying Comparison on UAV image datasets. $\uparrow$ indicates higher is better, while $\downarrow$ indicates lower is better. Best and second-best are highlighted in \textbf{bold} and \underline{underlined}. $\setminus$ denotes tracking failure due to low spatiotemporal continuity.}
\label{UAV_images_Tab}
\resizebox{\textwidth}{!}{
\begin{tabular}{cc|ccccccccc|c}

\toprule

\multicolumn{2}{c|}{Scene}                           & WHU            & NH1            & NH2            & NH3            & NH4            & npu            & phantom3       & ground         & park                              & \multirow{3}{*}{Avg.} \\
\multicolumn{2}{c|}{Image Number}                    & 242            & 132            & 158            & 153            & 163            & 300            & 399            & 295            & 300                               &                       \\
\multicolumn{2}{c|}{Pose Estimation (s)}             & 526.09         & 223.723        & 359.09         & 318.75         & 319.61         & 517.24         & 676.27         & 641.30         & 652.17                            &                       \\ \midrule
\multirow{5}{*}{AdR-Gaussian (7K)} & PSNR (dB)$\uparrow$& 24.99          & 25.71          & 24.28          & \underline{24.44}          & 25.61          & 26.85          & 26.83          & 25.44          & 25.59                             & 25.53                 \\
                                   & SSIM$\uparrow$  & \underline{0.82}     & \underline{0.83}     & \underline{0.81}     & \underline{0.82}     & 0.85           & \underline{0.81}     & 0.83           & 0.91           & 0.89                              & 0.84                  \\
                                   & LPIPS$\downarrow$& 0.17           & \underline{0.28}     & 0.27           & \underline{0.28}     & 0.23           & 0.29           & 0.26           & 0.27           & 0.35                              & 0.27                  \\
                                   & Model Size (MB)$\downarrow$& 839.2          & 346.2          & 299.8          & 286.1          & 277.1          & 895.5          & 809.1          & 754.1          & 581.2                             & 565.4                 \\
                                   & Cost Time (s)$\downarrow$& 794.18         & 432.47         & 561.51         & 521.26         & 515.80         & 786.22         & 1004.60        & 932.73         & 930.12                            & 719.88                \\ \midrule
\multirow{5}{*}{GS-SLAM}           & PSNR (dB)$\uparrow$& 20.01          & 18.21          & 16.19          & 18.16          & 17.79          & 21.48          & 20.57          & 20.99          & 21.02                             & 19.38                 \\
                                   & SSIM$\uparrow$  & 0.48           & 0.47           & 0.45           & 0.49           & 0.53           & 0.65           & 0.66           & 0.55           & 0.55                              & 0.54                  \\
                                   & LPIPS$\downarrow$& 0.58           & 0.79           & 0.82           & 0.82           & 0.69           & 0.68           & 0.75           & 0.62           & 0.68                              & 0.71                  \\
                                   & Model Size (MB)$\downarrow$& \textbf{44.6}  & \textbf{6.3}   & \textbf{8.1}   & \textbf{8.3}   & \textbf{8.8}   & \textbf{38.2}  & \textbf{29.9}  & \textbf{33.1}  & \textbf{13.8}                     & \textbf{21.2}         \\
                                   & FPS$\uparrow$   & 0.23           & 0.25           & 0.26           & 0.27           & 0.26           & 0.21           & 0.26           & \underline{0.24}     & 0.28                              & 0.25                  \\ \midrule
\multirow{5}{*}{On-the-Fly NVS}    & PSNR (dB)$\uparrow$& 19.64          & 18.12          & \textbf{24.96} & 17.81          & 22.01          & 21.54          & \multicolumn{3}{c|}{\multirow{5}{*}{\textbackslash{}}}               & 20.68                 \\
                                   & SSIM$\uparrow$  & 0.51           & 0.57           & 0.79           & 0.59           & 0.68           & 0.66           & \multicolumn{3}{c|}{}                                                & 0.63                  \\
                                   & LPIPS$\downarrow$& 0.47           & 0.43           & \underline{0.22}     & 0.42           & 0.33           & 0.67           & \multicolumn{3}{c|}{}                                                & 0.42                  \\
                                   & Model Size (MB)$\downarrow$& \underline{196.2}    & 218.7          & \underline{190.4}    & 179.8          & 121.7          & \underline{220.1}    & \multicolumn{3}{c|}{}                                                & \underline{187.8}           \\
                                   & FPS$\uparrow$   & \textbf{2.43}  & \textbf{2.81}  & \textbf{2.51}  & \textbf{2.56}  & \textbf{2.23}  & \textbf{2.33}  & \multicolumn{3}{c|}{}                                                & \textbf{2.48}         \\ \midrule
\multirow{5}{*}{On-the-Fly GS}     & PSNR (dB)$\uparrow$& \textbf{27.93} & \textbf{27.09} & 23.79          & 23.24    & \underline{27.27}    & \textbf{33.75} & \underline{31.46}    & \textbf{28.09} & \multirow{5}{*}{\textbackslash{}} & \underline{27.83}           \\
                                   & SSIM$\uparrow$  & \textbf{0.87}  & \textbf{0.86}  & 0.80           & 0.79           & \underline{0.87}     & \textbf{0.96}  & \underline{0.92}     & \underline{0.93}     &                                   & \underline{0.88}            \\
                                   & LPIPS$\downarrow$& \textbf{0.11}  & \textbf{0.23}  & \textbf{0.21}  & 0.29           & \underline{0.19}     & \underline{0.16}     & \textbf{0.12}  & \underline{0.22}     &                                   & \textbf{0.19}         \\
                                   & Model Size (MB)$\downarrow$& 534.4          & 403.2          & 352.2          & 385.1          & 324.3          & 672.3          & 831.9          & 692.9          &                                   & 524.5                 \\
                                   & FPS$\uparrow$   & 0.30           & 0.42           & 0.34           & 0.34           & 0.33           & 0.37           & \underline{0.34}     & \underline{0.24}     &                                   & 0.34                  \\ \midrule
\multirow{5}{*}{Ours}              & PSNR (dB)$\uparrow$& \underline{27.41}    & \underline{26.57}    & \underline{24.33}    & \textbf{24.51} & \textbf{27.28} & \underline{33.46}    & \textbf{31.81} & \underline{27.97}    & \textbf{27.42}                    & \textbf{27.86}        \\
                                   & SSIM$\uparrow$  & \textbf{0.87}  & \textbf{0.86}  & \textbf{0.82}  & \textbf{0.83}  & \textbf{0.88}  & \textbf{0.96}  & \textbf{0.94}  & \textbf{0.94}  & \textbf{0.93}                     & \textbf{0.89}         \\
                                   & LPIPS$\downarrow$& \underline{0.16}     & \textbf{0.23}  & 0.23           & \textbf{0.24}  & \textbf{0.17}  & \textbf{0.15}  & \underline{0.16}     & \textbf{0.19}  & \textbf{0.22}                     & \textbf{0.19}         \\
                                   & Model Size (MB)$\downarrow$& 230.4          & \underline{123.5}    & 204.7          & \underline{172.1}    & \underline{115.6}    & 265.8          & \underline{182.5}    & \underline{243.8}    & \underline{328.2}                       & 207.4                 \\
                                   & FPS$\uparrow$   & \underline{0.46}     & \underline{0.59}     & \underline{0.44}     & \underline{0.48}     & \underline{0.51}     & \underline{0.58}     & \textbf{0.59}  & \textbf{0.46}  & \textbf{0.46}                     & \underline{0.51}           \\

\bottomrule

\end{tabular}
}
\end{table*}

AdR-Gaussian and On-the-Fly GS achieve stable and good rendering results across all scenes. However, compared with A-TDOM, AdR-Gaussian exhibits noticeably inferior rendering quality on all scenes and requires significantly higher memory consumption as well as longer training time. Although On-the-Fly GS achieves rendering quality comparable to A-TDOM, it results in substantially larger model sizes and lower training efficiency. Moreover, due to GPU memory limitations, it also fails to complete the 3DGS training on the \textit{park} scene. In contrast, A-TDOM successfully completes 3DGS training on all scenes while exhibiting more stable rendering quality and smaller model size. These results underscore the advantages of A-TDOM when handling large-scale and complex scenes, demonstrating the strong potential for near real-time deployment in complex scenarios compared with existing SOTA approaches.

\subsection{\justifying Performance on Active TDOM Generation}

\begin{figure*}[ht]
    \centering
    \includegraphics[width=\textwidth]{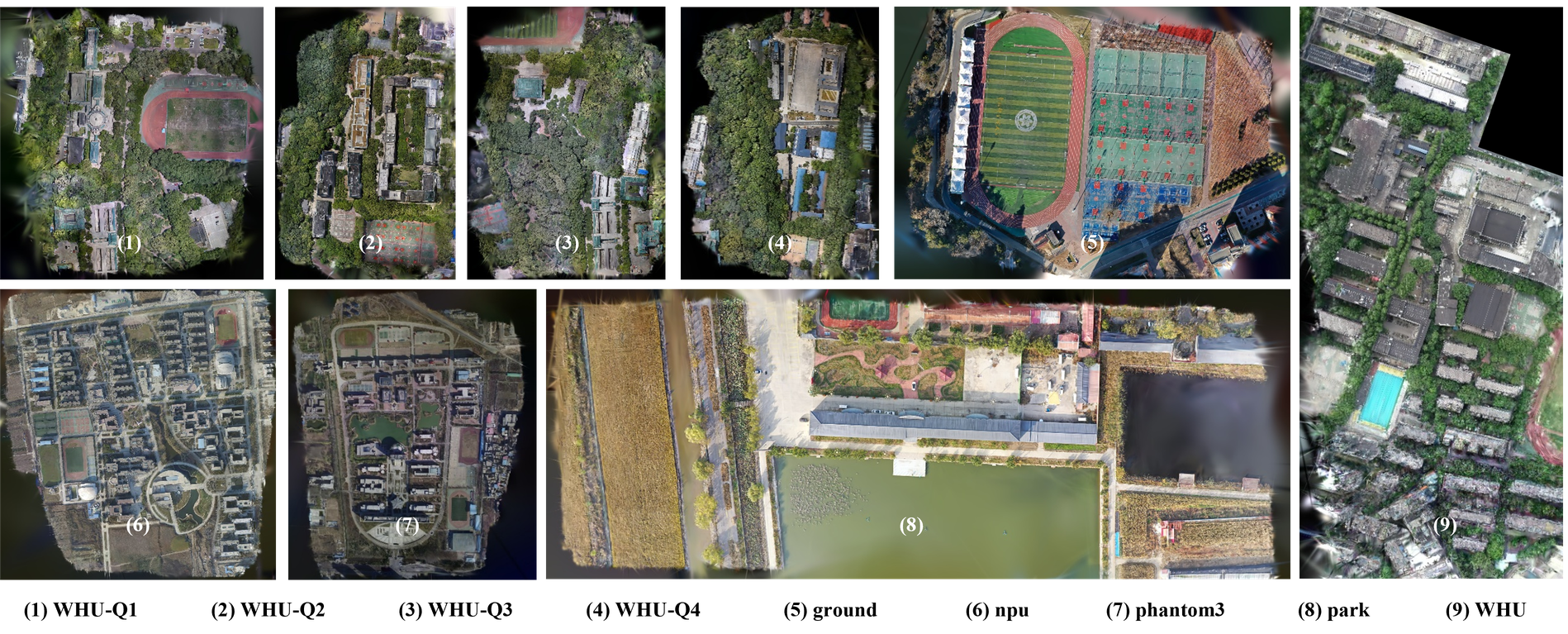}
    \caption{The final overall TDOMs of all UAV datasets using our A-TDOM.}
    \label{TDOM_Rendering_Ours}
\end{figure*}

\begin{figure*}[ht]
    \centering
    \includegraphics[width=\textwidth]{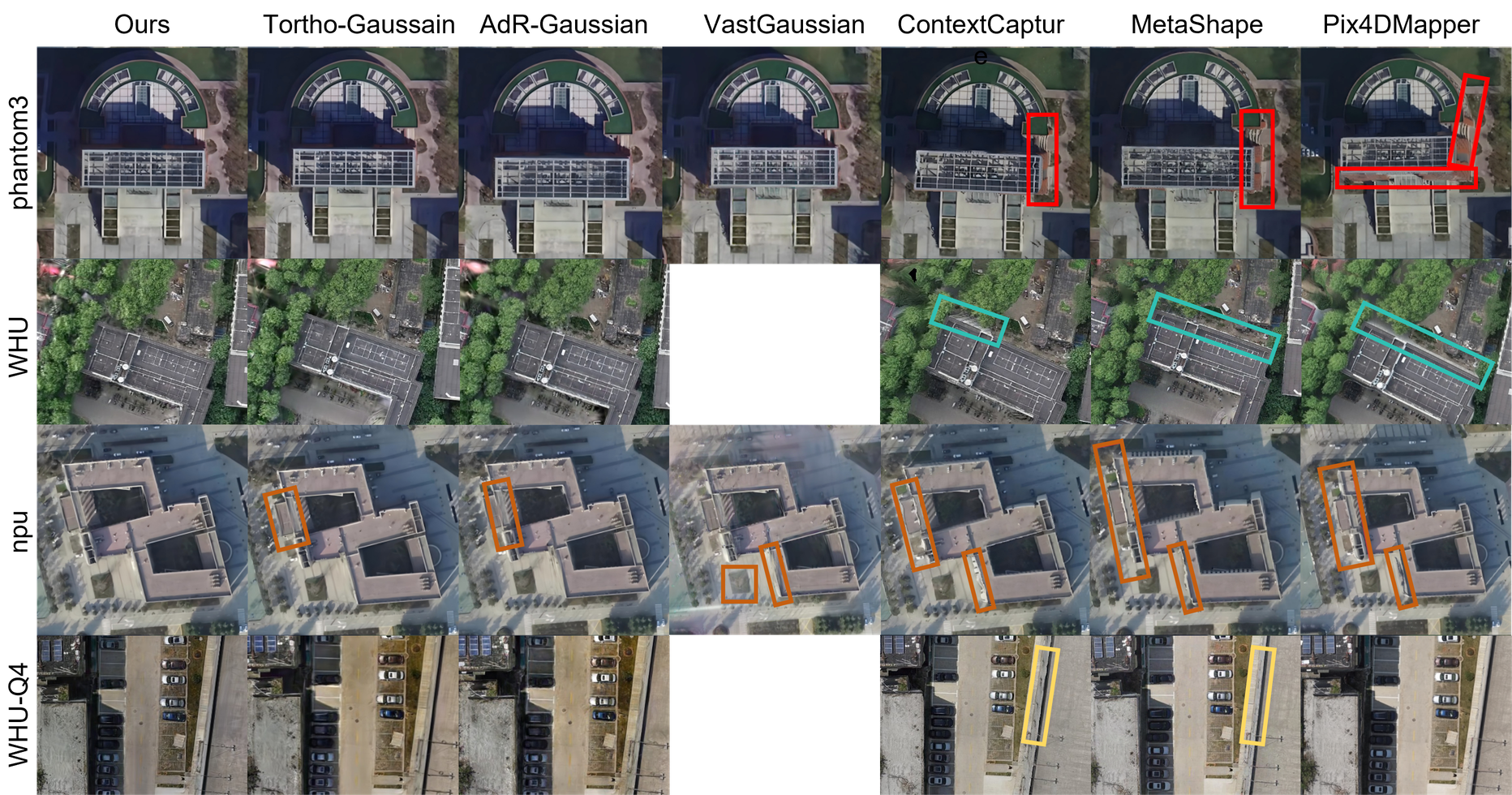}
    \caption{Local visual comparisons of the TDOM generated by different methods. Our A-TDOM achieves better building façade removal and exhibits visual quality that is comparable to other methods. While the commercial software generally demonstrate higher visual quality, they still suffer from incomplete façade removal in certain cases. Blank indicates 3DGS training failure. Colored boxes highlight regions where building façades are not fully removed or where blurring artifacts occur.}
    \label{TDOM_Qualitative}
\end{figure*}

To evaluate the performance of actively generating TDOM during image capturing, we select the nine UAV datasets specifically designed for large-scale scene reconstruction. Fig. \ref{TDOM_Rendering_Ours} presents the final overall TDOM generated by our A-TDOM, it implies that the tested datasets encompass a variety of scenes, ranging from residential buildings to diverse ground objects with rich textural details, thereby enabling a comprehensive qualitative and quantitative assessment of facade suppression efficacy, overall reconstruction fidelity and geometric accuracy.

In this study, the proposed A-TDOM is compared against three 3DGS-based methods (AdR-Gaussian, Tortho-Gaussian, and VastGaussian) and three commercial software (ContextCapture, MetaShape, and Pix4DMapper). Notably, AdR-Gaussian, Tortho-Gaussian, and VastGaussian are offline 3DGS optimization frameworks, which leverage the SfM results obtained from the On-the-Fly SfM pipeline after all images have been processed. Subsequently, each method performs 3DGS optimization for 30K iterations using the default parameter configurations prescribed in their original implementations. Since neither VastGaussian nor AdR-Gaussian incorporates a dedicated TDOM generation module, their TDOM outputs are produced by adopting the same orthorectification and projection strategy employed in Tortho-Gaussian~\citep{Tortho-gaussian}. For the commercial software, only the raw image inputs are provided, and TDOMs are generated using each platform’s default processing settings without additional manual intervention.

\begin{table*}[ht]
\centering
\caption{\justifying Comparison on 3DGS-based methods. \textbf{Ours\_Part} reports reports the results of the proposed A-TDOM method restricted to the subset of scenes on which VastGaussian successfully completes both 3DGS training and subsequent TDOM generation. $\uparrow$ indicates higher is better, while $\downarrow$ indicates lower is better. Best and second-best are highlighted in \textbf{bold} and \underline{underlined}. Furthermore, metric values marked in \textcolor{red}{(red)} indicate worse performance relative to our method, whereas those in \textcolor{blue}{(blue)} denote better performance. Additionally, values reported in parentheses represent the slowdown factor and model size ratio, respectively, i.e., how many times slower the method is and how much larger its model size is compared to our A-TDOM.}
\label{TDOM_Rendering_ModelSize}
\resizebox{0.75\textwidth}{!}{
\begin{tabular}{ccccccc}

\toprule


Metrics                                                         & PSNR (dB)$\uparrow$                                              & SSIM$\uparrow$                                         & LPIPS$\downarrow$                                               & Model Size (MB)$\downarrow$                              & Cost Time (s)$\downarrow$                              \\ \midrule
\begin{tabular}[c]{@{}c@{}}AdR-Gaussian\\ (7K)\end{tabular}     & \begin{tabular}[c]{@{}c@{}}23.31\\ \textcolor{red}{(-2.84)}\end{tabular}          & \begin{tabular}[c]{@{}c@{}}0.76\\ \textcolor{red}{(-0.10)}\end{tabular} & \begin{tabular}[c]{@{}c@{}}0.38\\ \textcolor{red}{(+0.14)}\end{tabular}          & \begin{tabular}[c]{@{}c@{}}619.4\\ (×2.65)\end{tabular}  & \begin{tabular}[c]{@{}c@{}}\underline{757}\\ (×1.48)\end{tabular}  \\
\begin{tabular}[c]{@{}c@{}}AdR-Gaussian\\ (30K)\end{tabular}    & \begin{tabular}[c]{@{}c@{}}\textbf{27.71}\\ \textcolor{blue}{(+1.56)}\end{tabular} & \begin{tabular}[c]{@{}c@{}}\underline{0.80}\\ \textcolor{red}{(-0.06)}\end{tabular} & \begin{tabular}[c]{@{}c@{}}\textbf{0.23}\\ \textcolor{blue}{(-0.01)}\end{tabular} & \begin{tabular}[c]{@{}c@{}}1591.1\\ (×6.81)\end{tabular} & \begin{tabular}[c]{@{}c@{}}2333\\ (×4.56)\end{tabular} \\
\begin{tabular}[c]{@{}c@{}}Tortho-Gaussian\\ (30K)\end{tabular} &\begin{tabular}[c]{@{}c@{}}\underline{26.67}\\ \textcolor{blue}{(+0.52)}\end{tabular}&\begin{tabular}[c]{@{}c@{}}0.73\\ \textcolor{red}{(-0.13)}\end{tabular} & \begin{tabular}[c]{@{}c@{}}0.31\\ \textcolor{red}{(+0.07)}\end{tabular}          & \begin{tabular}[c]{@{}c@{}}\underline{344.7}\\ (×1.48)\end{tabular}  & \begin{tabular}[c]{@{}c@{}}1438\\ (×2.81)\end{tabular} \\
Ours                                                            & 26.15                                                            & \textbf{0.86}                                          & \underline{0.24}                                                & \textbf{233.6}                                           & \textbf{512}                                           \\ \midrule \midrule
\begin{tabular}[c]{@{}c@{}}VastGaussian\\ (30K)\end{tabular}    & \begin{tabular}[c]{@{}c@{}}\textbf{30.36}\\ \textcolor{blue}{(+0.19)}\end{tabular} & \begin{tabular}[c]{@{}c@{}}\underline{0.90}\\ \textcolor{red}{(-0.04)}\end{tabular} & \begin{tabular}[c]{@{}c@{}}\textbf{0.10}\\ \textcolor{blue}{(-0.08)}\end{tabular} & \begin{tabular}[c]{@{}c@{}}\underline{1104.3}\\ (×4.33)\end{tabular} & \begin{tabular}[c]{@{}c@{}}\underline{5887}\\ (×9.29)\end{tabular} \\
Ours\_Part                                                      & \underline{30.17}                                                & \textbf{0.94}                                          & \underline{0.18}                                                & \textbf{255.1}                                           & \textbf{634}                                            \\

\bottomrule

\end{tabular}
}
\end{table*}

\textbf{Qualitative Experiments.} In the qualitative evaluation, we primarily compare the effectiveness of different methods in building façades removal and the visual quality of the generated TDOM. Fig. \ref{TDOM_Qualitative} presents local visual comparisons of the TDOM generated by different methods.

From this visual comparison, we can see that our A-TDOM and the other three 3DGS-based methods effectively eliminate building façades, whereas the TDOM generated by commercial software solutions still retain a certain of residual façade structures. This indicates that 3DGS-based methods can achieve superior orthographic projection results using a comparatively simpler algorithmic pipeline compared to traditional TDOM generation workflows. Regarding visual quality of the generated TDOM, our A-TDOM achieves rendering quality which is slightly inferior to that of offline 3DGS training solutions or commercial software. Moreover, A-TDOM is capable of producing visually consistent and high-quality results in texture-rich regions. These results sufficiently demonstrate the feasibility and effectiveness of our proposed method for TDOM generation.

\begin{table*}[ht]
\centering
\caption{\justifying Cost Time (s) of TDOM generation. \textbf{Ours\_Part} reports the results of A-TDOM only on scenes where VastGaussian successfully completes 3DGS training and subsequent TDOM generation. \textbf{Ours\_FPS} is the averaging frames per second of our A-TDOM for generating TDOM. For clarity, the best and second-best results are highlighted in \textbf{bold} and \underline{underlined}, respectively. Additionally, values in parentheses indicate the slowdown factor, i.e., how many times slower other methods are relative to our A-TDOM.}
\label{TDOM_Time}
\resizebox{\textwidth}{!}{
\begin{tabular}{ccccccccccccc}

\toprule

Scene                                                           & WHU              & WHU\_Q1          & WHU\_Q2          & WHU\_Q3          & WHU\_Q4          & npu          & phantom3     & ground       & park         & Avg.                                                     \\ \midrule
ContextCapture                                                  & 2993             & 1709             & 1218             & 1102             & 1164             & 1240         & 1606         & 1326         & 1277         & \begin{tabular}[c]{@{}c@{}}1515\\ (×2.96)\end{tabular}   \\
MetaShape                                                       & 1231             & 1744             & 880              & 763              & 758              & 2533         & 2583         & 1531         & 1483         & \begin{tabular}[c]{@{}c@{}}1501\\ (×2.93)\end{tabular}   \\
Pix4DMapper                                                     & 816              & \underline{487}  & \underline{540}  & \underline{488}  & \underline{504}  & 1601         & 1443         & 1386         & 1448         & \begin{tabular}[c]{@{}c@{}}968 \\ (×1.89)\end{tabular}   \\ \midrule
\begin{tabular}[c]{@{}c@{}}AdR-Gaussian\\ (7K)\end{tabular}     & \underline{794}  & 578              & 644              & 558              & 587              &\underline{786}&\underline{1005}&\underline{933}&\underline{930}&\begin{tabular}[c]{@{}c@{}}\underline{757}\\ (×1.48)\end{tabular}\\
\begin{tabular}[c]{@{}c@{}}AdR-Gaussian\\ (30K)\end{tabular}    & 2231             & 2071             & 2251             & 1834             & 2366             & 2321         & 2585         & 2808         & 2529         & \begin{tabular}[c]{@{}c@{}}2333\\ (×4.56)\end{tabular}   \\
\begin{tabular}[c]{@{}c@{}}Tortho-Gaussian\\ (30K)\end{tabular} & 1295             & 1049             & 1480             & 1256             & 1371             & 1391         & 1574         & 1610         & 1913         & \begin{tabular}[c]{@{}c@{}}1438\\ (×2.81)\end{tabular}   \\
\begin{tabular}[c]{@{}c@{}}VastGaussian\\ (30K)\end{tabular}    & \textbackslash{} & \textbackslash{} & \textbackslash{} & \textbackslash{} & \textbackslash{} & 5623         & 5681         & 6283         & 5961         & \begin{tabular}[c]{@{}c@{}}5887\\ (×9.29)\end{tabular}   \\ \midrule
Ours/Ours\_Part                                                 & \textbf{526}     & \textbf{370}     & \textbf{425}     & \textbf{374}     & \textbf{381}     & \textbf{517} & \textbf{725} & \textbf{641} & \textbf{652} & \textbf{512/634}                                         \\
Ours\_FPS                                                       & 0.46             & 0.67             & 0.56             & 0.66             & 0.61             & 0.58         & 0.59         & 0.46         & 0.46         & 0.56                                                     \\

\bottomrule

\end{tabular}
}
\end{table*}

\textbf{Quantitative Experiments.} In this section, we first investigate the time cost required for TDOM generation by different methods. For our A-TDOM, the reported cost time spans from the acquisition of the first input image to the completion of 3DGS training and TDOM generation. For the offline 3DGS-based methods, since they start 3DGS training only after the On-the-Fly SfM has processed all input images, the total time cost is computed as the sum of the time consumed by On-the-Fly SfM, and the time for completing 3DGS training and generating the TDOM. Notably, although VastGaussian divides scene into several blocks and supports parallel training, all experiments are conducted on a single NVIDIA RTX 4090 GPU. When parallel training of all blocks is infeasible due to VRAM limitations, multiple runs are performed and the accumulated runtime across all runs is reported as the total time cost of VastGaussian. For the commercial software, the elapsed time from the start of image processing to the acquisition of the TDOM is measured as the time cost. In addition to runtime, the metrics of rendering quality and the model size of each 3DGS field are also used for the evaluation. The quantitative results are summarized in Tab. \ref{TDOM_Rendering_ModelSize} and Tab. \ref{TDOM_Time}, where Tab. \ref{TDOM_Rendering_ModelSize} presents the average rendering quality and the average model size of A-TDOM and other 3DGS-based methods over all scenes and Tab. \ref{TDOM_Time} lists the time consumption of different methods for TDOM generation across various scenes. 


The experimental results presented in the two tables demonstrate that our proposed A-TDOM method exhibits significant advantages in time efficiency, consistently requiring substantially less time to generate TDOMs across all evaluated scenes compared to all competing approaches. When compared to AdR-Gaussian trained for 7K iterations, A-TDOM achieves notable improvements not only in rendering quality but also in runtime efficiency and model compactness. In contrast to both AdR-Gaussian and Tortho-Gaussian trained for 30K iterations, A-TDOM incurs only a moderate degradation in rendering fidelity while delivering substantial gains in both computational speed and model size reduction. For VastGaussian, it fails to complete 3D Gaussian Splatting (3DGS) training and TDOM generation for the \textit{WHU\_Q1}, \textit{Q2}, \textit{Q3} and \textit{Q4} scenes due to GPU memory (VRAM) exhaustion. Moreover, in the case of the full WHU scene, VastGaussian’s partitioning strategy breaks down, resulting in severe structural omissions in the reconstructed 3DGS model. Consequently, this instance is also categorized as a failure. For the remaining scenes where VastGaussian successfully produces a TDOM, A-TDOM still demonstrates competitive performance: although its rendering quality is marginally lower, VastGaussian exhibits the highest computational time among all baselines and produces models with significantly larger storage footprints.

Collectively, these findings indicate that, relative to existing 3DGS-based methods, A-TDOM offers a highly favorable trade-off, dramatically reducing both time consumption and model size, while compromising only minimally on rendering quality. 

Comparing to the commercial software, ContextCapture, MetaShape, and Pix4DMapper need ×2.96, ×2.93, and ×1.89 more cost time than A-TDOM, respectively. Therefore, our A-TDOM is capable of rapid TDOM generation while maintaining the degradation in quality within an acceptable range. Furthermore, our A-TDOM maintains an overall processing speed exceeding 0.5 FPS, allowing the TDOM to be updated within 1–2 seconds following the acquisition of each new image. This demonstrates that A-TDOM effectively enables near real-time active TDOM generation.

\begin{table*}[ht]
\centering
\caption{\justifying Results of RMSE from six GCPs in the npu subset. $\downarrow$ indicates lower is better. Best and second-best are highlighted in \textbf{bold} and \underline{underlined}, respectively.}
\label{table:gcp_rmse_comparison}
\resizebox{\textwidth}{!}{
\begin{tabular}{lccccccc}
\toprule
\textbf{Method} & ContextCapture & Metashape & Pix4DMapper & AdR-Gaussian& Tortho-Gaussian & VastGaussian& Ours\\
\midrule
\textbf{RMSE (m)}$\downarrow$ & 3.130 & 2.530 & \textbf{1.502} & 2.111 & 1.920 & 1.922 & \underline{1.780} \\
\bottomrule
\end{tabular}
}
\end{table*}

\begin{figure*}[ht]
    \centering
    \includegraphics[width=0.95\textwidth]{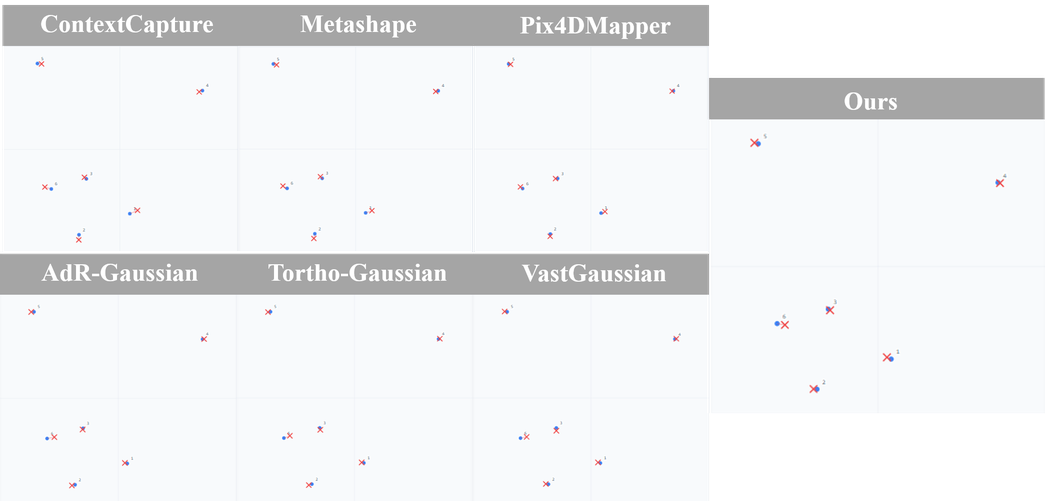}
    \caption{Relative spatial relationship between the adjusted GCPs and their corresponding 2D positions on the TDOMs. The blue dots represent the GCP positions projected onto the XY-plane, while the red crosses denote the aligned pixel positions mapped into the real-world coordinate system.}
    \label{visualRMSE}
\end{figure*}

To ensure a rigorous quantitative evaluation on the geometric accuracy of TDOM generated by different methods, we utilize the six ground control points (GCPs) provided in the NPU dataset$\footnote{See more details of the NPU datasets at \href{ https://www.adv-ci.com/blog/source/npu-drone-map-dataset/}{ https://www.adv-ci.com/blog/source/npu-drone-map-dataset/}}$. First, the GCPs, originally given in geodetic coordinates, are transformed into a local East-North-Up (ENU) coordinate system. A minor rotational correction is then applied to align the Z-axis perpendicular to the TDOM plane, ensuring consistency with the orthographic projection assumption. Next, we extract the XY-plane and manually label the corresponding GCP locations in each TDOM to obtain their 2D pixel coordinates. A 2D registration is performed using least-squares adjustment between the two planes' coordinates. The RMSE across all six GCPs is then computed, as summarized in Tab. \ref{table:gcp_rmse_comparison}. Our proposed A-TDOM method achieves an RMSE of 1.780 meters, ranking second among all evaluated approaches and only slightly lower than Pix4DMapper. As illustrated in Fig.~\ref{visualRMSE}, the spatial correspondence between the GCPs and their registered 2D points on the TDOMs further corroborates these quantitative findings: Both A-TDOM and Pix4DMapper exhibit notably tighter alignment, with minimal positional deviation. This demonstrates that our method achieves geometric accuracy comparable to the leading commercial photogrammetry software, making it a viable solution for high-precision mapping applications.

\subsection{Ablation Studies}

\begin{figure}[ht]
    \centering
    \includegraphics[width=0.45\textwidth]{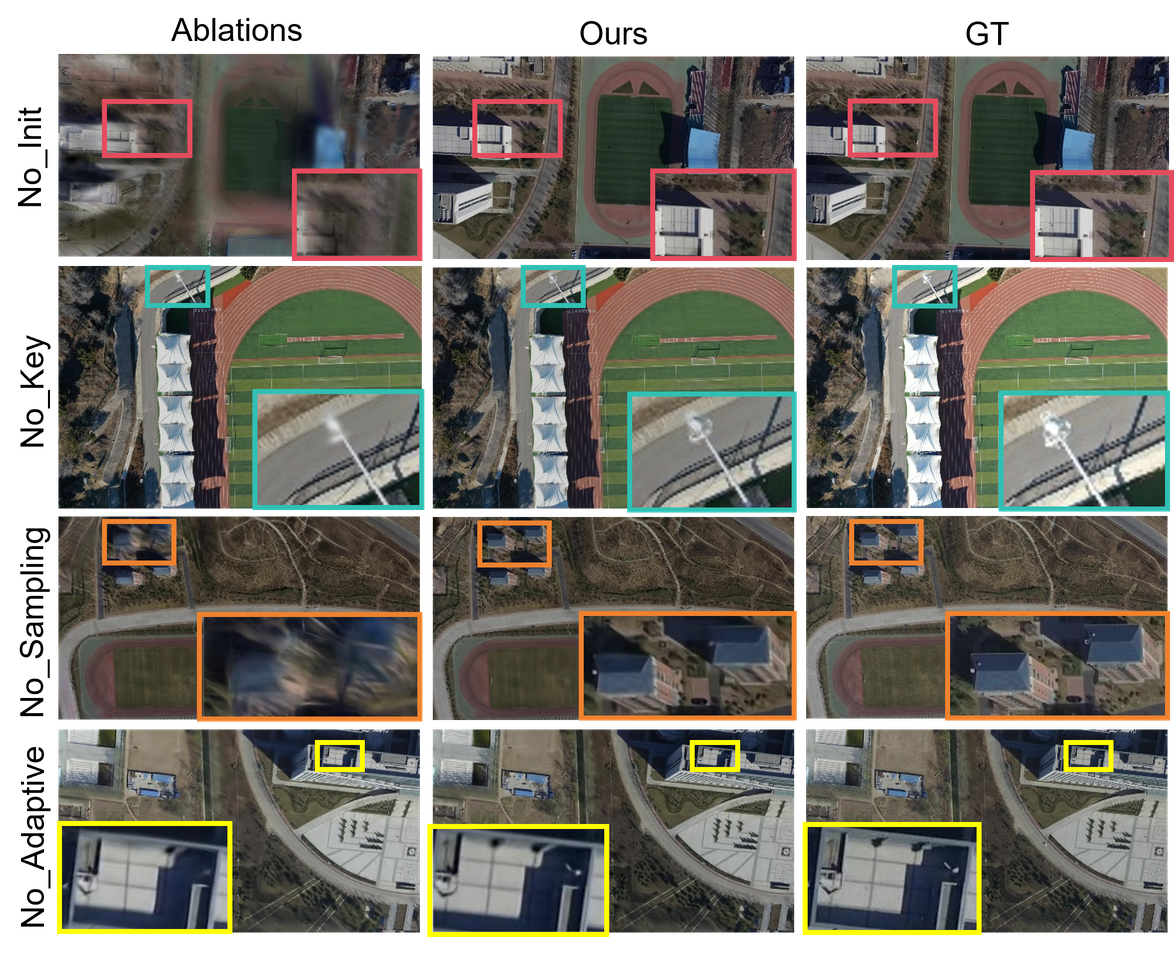}
    \caption{Ablation Studies of rendering results. This indicate that each component substantially contributes to the 3DGS training and rendering performance of A-TDOM.}
    \label{Ablation_Fig}
\end{figure}

\begin{table*}[ht]
\centering
\caption{\justifying Ablation experiment of the proposal components in the A-TDOM method. $\uparrow$ indicates higher is better, while $\downarrow$ indicates lower is better. Best and second-best are highlighted in \textbf{bold} and \underline{underlined}, respectively. $\setminus$ denotes training failure due to VRAM limitation.}
\label{Ablation_Tab}
\resizebox{0.75\textwidth}{!}{
\begin{tabular}{ccccccc}

\toprule

Scene                     & Methods      & FPS$\uparrow$ & Model Size (MB)$\downarrow$& PSNR$\uparrow$& SSIM$\uparrow$& LPIPS$\downarrow$\\ \midrule
\multirow{5}{*}{phantom3} & No\_Init     & 0.54          & \underline{183.2}     & 30.83          & 0.93          & 0.19          \\
                          & No\_Key      & \underline{0.56}    & 192.4           & 31.22          & 0.93          & 0.19          \\
                          & No\_Sampling & 0.21          & 990.8           & 31.21          & 0.94          & \textbf{0.15} \\
                          & No\_Adaptive & 0.41          & 293.1           & \textbf{32.14} & \textbf{0.95} & 0.17          \\
                          & Ours         & \textbf{0.59} & \textbf{182.5}  & \underline{31.81}    & \underline{0.94}    & \underline{0.16}    \\ \midrule \midrule
\multirow{5}{*}{ground}   & No\_Init     & 0.39             & 290.9            & 26.89            & 0.92             & 0.22             \\
                          & No\_Key      & \underline{0.42}       & \underline{257.7}      & 27.34            & \underline{0.93}       & 0.21             \\
                          & No\_Sampling & \textbackslash{} & \textbackslash{} & \textbackslash{} & \textbackslash{} & \textbackslash{} \\
                          & No\_Adaptive & 0.34             & 448.4            & \textbf{28.46}   & \textbf{0.94}    & \textbf{0.18}    \\
                          & Ours         & \textbf{0.46}    & \textbf{243.8}   & \underline{27.97}& \textbf{0.94}    & \underline{0.19} \\
\bottomrule

\end{tabular}
}
\end{table*}

Ablation studies are performed to validate the efficacy of each individual innovative component in this work. Based on the UAV scene - \textit{phantom3} and \textit{ground}, all ablation results are summarized in Tab. \ref{Ablation_Tab} and Fig. \ref{Ablation_Fig}. First, the threshold which sets the number of already oriented images before starting the 3DGS estimation is set to 3, which is the minimum images for On-the-Fly SfM (No\_init). Second, we disable the key-region extraction step, meaning that regions with insufficient image overlap are included in the training (No\_Key). Third, we use the densification strategy from the original 3DGS implementation instead of our Gaussian Sampling and Integration strategy (No\_Sampling). Finally, we remove the proposed Adaptive 3DGS Optimization Strategies and adopt the original offline training scheme of 3DGS, in which iterations are uniformly distributed among all images, and the learning-rate decay is scheduled based on the global training progress of the 3DGS field (No\_Adaptive).

In our A-TDOM framework, initializing the 3DGS field with only three input images (No\_Init) leads to severe overfitting to these early views. This overfitting biases the initial 3DGS representation, resulting in a poorly conditioned density field that compromises the effectiveness of subsequent optimization. Although regions distant from the initial 3DGS field are also affected, the overfitting induces significant geometric and appearance distortions in the vicinity of the initial views, severely degrading local rendering fidelity and, consequently, the overall quality of the reconstructed scene.

In the absence of key-region extraction (No\_Key), the optimization process treats all regions uniformly, regardless of whether they exhibit sufficient multi-view overlap. As a result, additional Gaussians are unnecessarily allocated to poorly constrained areas with sparse or inconsistent observations. This not only inflates the model size but also undermines rendering efficiency. In contrast, our key-region extraction mechanism provides explicit spatial guidance, directing Gaussian allocation toward well-constrained, semantically or geometrically salient regions. This targeted optimization enables A-TDOM to reconstruct fine-grained scene details using only the necessary Gaussians, thereby avoiding redundancy. Without this mechanism, the model accumulates superfluous Gaussians, leading to increased memory footprint, slower training convergence, and degraded rendering quality.

Replacing the proposed Gaussian Sampling and Integration strategy with the standard densification scheme from conventional 3DGS (No\_Sampling) results in a substantial increase in model size and degradation in training efficiency, without any measurable gain in rendering quality. The dramatic expansion of the model size even exceeds the available VRAM limitation on the UAV image dataset \textit{ground}, causing 3DGS training failure. This ablation underscores the efficacy of our sampling and integration strategy, which enables more precise and parsimonious Gaussian insertion, maintaining high visual fidelity while minimizing redundancy. 

Finally, adopting the training strategy used in offline 3DGS training solutions (No\_Adaptive) yields only a slight improvement in rendering quality at the cost of significantly larger model size and poorer training efficiency. In this setting, training iterations are evenly distributed across all registered image,  which impedes rapid adaptation to newly acquired views. Consequently, the optimizer compensates for insufficient per-image refinement by introducing additional Gaussians, further bloating the model. Moreover, the learning rate decay policy in offline 3DGS relies on a pre-specified total iteration count, an assumption incompatible with the streaming, open-ended nature of online reconstruction. Thus, such a strategy is ill-suited for real-time, incremental 3DGS optimization in our A-TDOM pipeline.

\section{Conclusions}\label{Conclusion}

In this paper, we proposed A-TDOM, an innovative workflow which enables simultaneous image acquisition and active TDOM generation based on On-the-Fly 3DGS. First, key regions for each input image are determined via reprojected Delaunay triangulation masking, which prevents reconstruction in low-overlap areas. Next, a novel Gaussian Sampling and Integration method is proposed for online 3DGS field expansion. The coarsely reconstructed or previously unseen regions are identified and new Gaussians are sampled within these regions. Subsequently, the sampled Gaussians are integrated into the current 3DGS field, followed by 3DGS optimization. Furthermore, several modifications are applied to the iteration allocation and learning rate update strategies, enabling rapid optimization for coarsely reconstructed regions and stable refinement for well-reconstructed regions. After the optimization for each input image, TDOM is generated or updated through orthographic splatting. The experimental results demonstrate that, in terms of 3DGS training, A-TDOM successfully achieves near real-time 3DGS training while maintaining considerable rendering quality. In particular, when processing UAV imagery, compared with SLAM-based methods \citep{GS-SLAM, RTG-SLAM} and other approaches that implement online 3DGS training \citep{On-the-Fly-GS, On-the-Fly-NVS}, A-TDOM exhibits significant advantages in rendering quality, model size, and training speed. Regarding TDOM generation, compared with commonly used commercial software, A-TDOM can produce TDOM with satisfactory accuracy while requiring considerably less processing time, demonstrating its potential for practical applications in real-world production workflows.


\bibliography{sn-bibliography}

\end{document}